\documentclass{article}

\usepackage{arxiv}

\usepackage[utf8]{inputenc} 
\usepackage[T1]{fontenc}    
\usepackage{hyperref}       
\usepackage{url}            
\usepackage{booktabs}       
\usepackage{amsfonts}       
\usepackage{nicefrac}       
\usepackage{microtype}      
\usepackage{lipsum}		
\usepackage{graphicx}
\usepackage{natbib}
\usepackage{doi}
\usepackage{algorithm,algcompatible,amsmath}
\usepackage[caption=false,font=normalsize,labelfont=sf,textfont=sf]{subfig}
\algnewcommand\INPUT{\item[\textbf{Input:}]}%
\algnewcommand\OUTPUT{\item[\textbf{Output:}]}%
\usepackage{subfig}

\title{Image Classifier Based Generative Method for Planar Antenna Design}

\author{ \hspace{1mm}{Yang~Zhong} \thanks{This project is sponsored by Meta Internship Program.} \\
	Department of ECE, Duke University\\
	Durham, NC 27708, USA \\
	\texttt{hustzy@outlook.com} \\
	\And
	\hspace{1mm}{Weiping~Dou} \\
	Reality Laboratory, Meta\\
	Sunnyvale, CA 94087, USA \\
	\And
	\hspace{1mm}{Andrew~Cohen}\\
	AI Research, Meta\\
	Sunnyvale, CA 94087, USA \\
	\And
	\hspace{1mm}{Dia'a~Bisharat} \\
	Reality Laboratory, Meta\\
	Sunnyvale, CA 94087, USA \\
	\And
    \hspace{1mm}{Yuandong~Tian}\\
	AI Research, Meta\\
	Sunnyvale, CA 94087, USA \\
	\And
	\hspace{1mm}{Jiang~Zhu} \\
	Reality Laboratory, Meta\\
	Sunnyvale, CA 94087, USA \\
	\And
    \hspace{1mm}{Qing~Huo~Liu} \\
	Department of ECE, Duke University\\
	Durham, NC 27708, USA \\
}

\date{}

\renewcommand{\shorttitle}{Antenna Design Using Fixed-Dimension Elements}

\hypersetup{
pdftitle={A template for the arxiv style},
pdfsubject={q-bio.NC, q-bio.QM},
pdfauthor={David S.~Hippocampus, Elias D.~Striatum},
pdfkeywords={First keyword, Second keyword, More},
}

\begin{document}
\maketitle

\begin{abstract}
	To extend the antenna design on printed circuit boards (PCBs) for more engineers of interest, we propose a simple method that models PCB antennas with a few basic components. By taking two separate steps to decide their geometric dimensions and positions, antenna prototypes can be facilitated with no experience required. Random sampling statistics relate to the quality of dimensions are used in selecting among dimension candidates. A novel image-based classifier using a convolutional neural network (CNN) is introduced to further determine the positions of these fixed-dimension components. Two examples from wearable products have been chosen to examine the entire workflow. Their final designs are realistic and their performance metrics are not inferior to the ones designed by experienced engineers.
\end{abstract}


\section{Introduction}
\label{SecIntro}

Designing antennas in the wireless consumer electronic industry is a technical challenge that requires not only many efforts in simulation and measurement, but also experience in developing initial prototypes. The antenna space and the surrounding environment keep changing within various products. A well-designed antenna that meets the target of one product may not work with another even though they might come from the same production line. Selecting an initial antenna type, a monopole, loop or inverted F, to start with is critical. In many cases, it depends on who is the antenna engineer working on this project. For a same project and given the same specifications, different antenna engineers might surprisingly come out unalike types of antenna designs just because of their personalized experience and taste. In this era of rapid product iterations, there is high demand of creative antenna designs and it is hard to find antenna expertise. Therefore, in this paper, we will present a workflow of proposing good prototypes that antenna design experience is not a mandatory requirement.

Antenna optimization have been widely studied and well presented in previous work, such as the trust region method \cite{Koziel2018ExpeditedDC}, particle swarm method \cite{Nanbo2007ParticleSO}, evolutionary strategies \cite{6612668} and many types of machine learning methods \cite{Sharma2020MachineLT,Koziel2021AccurateMO,Nan2021DesignOU,Naseri2021AGM,9670640,Abdullah2022SupervisedLearningBasedDO}. We can learn from these publications that their first and important step is to propose an initial prototype. For any successful antenna development, the difficulty of its optimization process and the quality of its electric and mechanical performance usually rely on the initial prototype that they selected. 

Geometric parameterization is the foundation of artificial-intelligence-assisted antenna design. In our previous work \cite{9887063}, we investigated antenna modeling with trapezoids-connected nodes for wired inverted F antenna with a small number of parameters. To make the antenna more practical for manufacturing, in this work, we decide to replace the nodes with conventional rectangles as the basic components. Unlike the literature where dimensions and positions are determined or optimized together, we intentionally separate the dimension and position parameters of the rectangles. This decomposition actually reduces the design complexity, whereas a special usage of random data statistics can avoid a bad local optimum.

When looking at well-designed antennas within the limited space of a compact wearable device, often five or six rectangles are sufficient to substantially decompose them. Therefore, the number of rectangles is predetermined; The next step is to determine the dimensions of these rectangles. For a random set of dimensional parameters, the prototype may never meet the target even after optimizing the positions of these fixed-dimension elements. Therefore, several sets of dimensions should be examined to avoid this situation happen. For each set, some samples with random positions are simulated in order to calculate the statistics of this set of dimensions. Such statistics indicate the likelihood of obtaining qualified antennas after optimizing the positions. 

\begin{figure}[!t]
\centering
{\includegraphics[width=1.0\columnwidth]{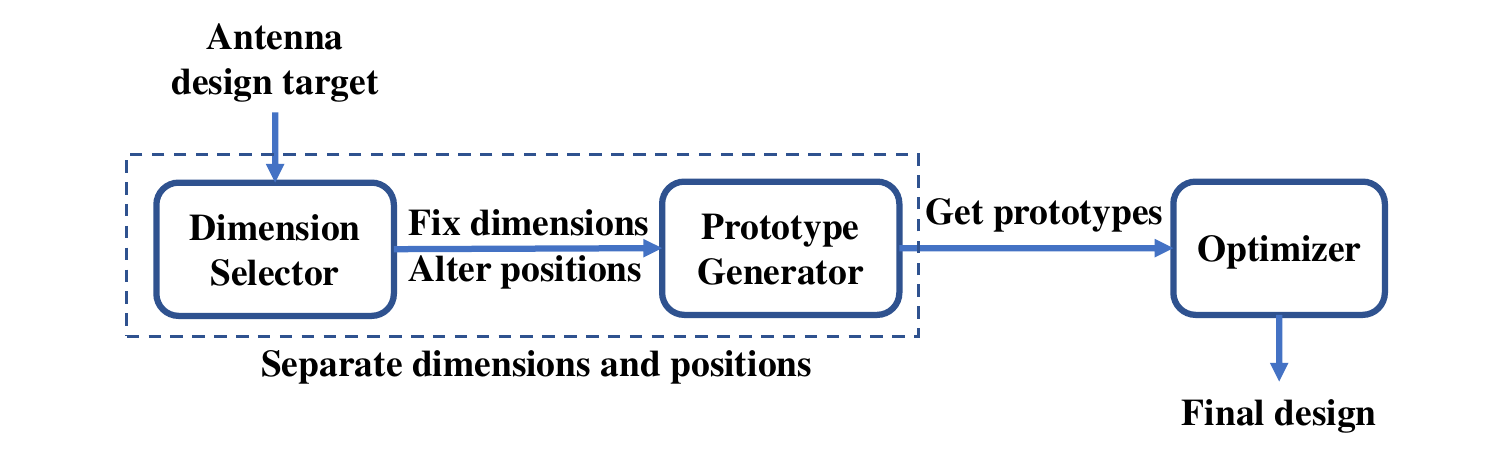}}
\caption{Overall workflow to design PCB antennas using basic components.}
\label{overall}
\end{figure}

Once the dimensions are selected, the initial antenna prototype can be finalized by changing the positions of these fixed-dimension rectangles. We continue to use the generative method in \cite{9906451} because we believe that it is a prospective route to automate antenna design. In this work, we use a random generator to avoid the overfitting issue, and an image-based filter for the discriminator. The core of this filter is a convolutional neural network that is trained by simulation data. The data are pairs of the 2D antenna geometric image and the simulated result. In order to reduce cost, we only simulate the antenna models that can pass the filter. The filter will be updated every time once a batch of simulation data are accumulated. After several iterations, the distribution of the simulated antennas will move towards the design target, which means that the opportunity of obtaining a better antenna prototype increases. The learnt prototype doesn't necessarily have to meet the design target because there are many optimizers \cite{ZHU2007TAP_Special_issue,Wu2020MultistageCM,wang2020learning,Zhou2021ATP} that can be used for further tuning. Figure.~\ref{overall} summarizes the antenna design workflow using fixed-dimension components.

This work introduces several novel contributions: 
\begin{itemize}
\item  Addressing two crucial geometric parameters, namely dimension and position, as distinct entities throughout the entire process of antenna prototype exploration.
\item Dimension selection is informed by the statistics of random placement, offering a unique approach to find a promising set of the fixed dimensions.
\item For components with fixed dimensions, the placement employs a generative method facilitated by an image classifier, providing an innovative means of achieving proper placements.
\end{itemize}
Therefore we review antenna design from layman's perspective, adopt the most popular image concept of artificial intelligence, and pursue the non-traditional antenna design approach to participate in antenna design. In the upcoming sections, we initially present the PCB antenna image classifier in Section \ref{classifier}, which also serves as the central element of the prototype generator. Subsequently, an example guides us through the process of dimension selection in Section \ref{dimension}, followed by position determination in Section \ref{position}. To further illustrate the concept, we present an additional practical antenna design example in Section \ref{real_example}. Concluding insights can be found in Section \ref{conclusion}.

\section{Image Classifier in Antenna Generative Method}
\label{classifier}

We present the image classifier originally centered on position determination separately at the beginning, as it can also find application in other antenna design methods. The geometric structures of most PCB antennas can be translated into images. When coupled with electromagnetic (EM) simulation results, these images form pairs of data. The dataset accumulated through simulations can then be used to train an image classifier. Once effectively trained, the classifier can predict, prior to EM simulations, whether a new antenna geometry model can achieve the desired EM performance goals.

\begin{figure}[!t]
\centering \includegraphics[width=1 \columnwidth]{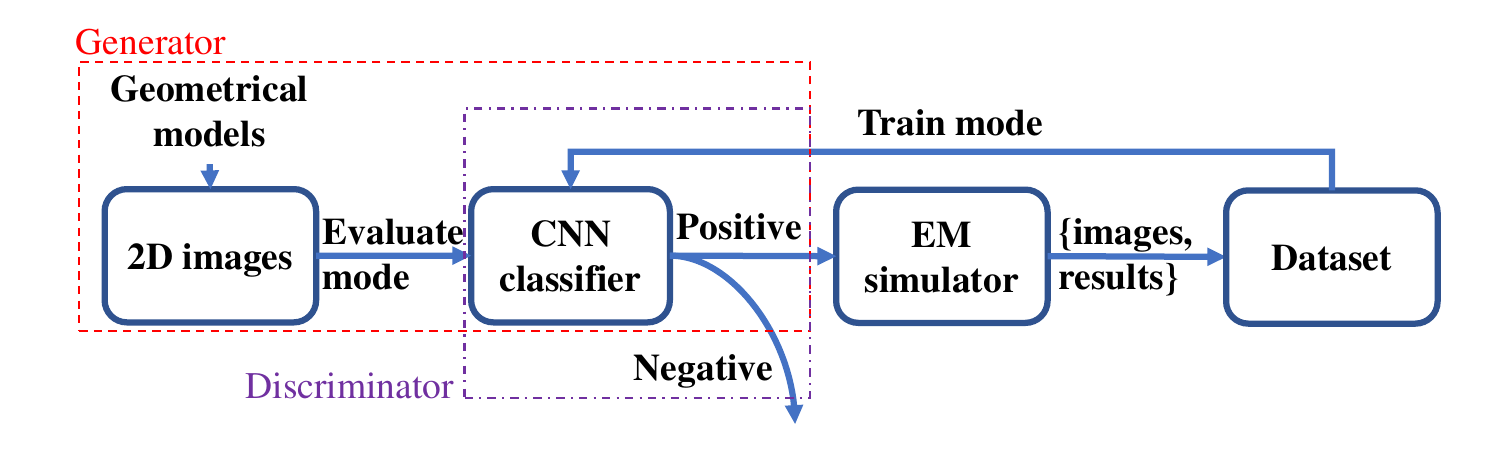}\\
  \caption{A CNN-driven image classifier functions as both a predictive score discriminator and a part of the PCB antenna prototype generator.}\label{classifier_in_generative}
\end{figure}

In light of large language models, achieving accurate and meaningful answers from the model necessitates a substantial number of parameters within the model itself. Consequently, the training dataset for the model must be extensive. However, for tasks such as antenna design, which involve intricate transformations of geometric and physical variables, the time consuming EM simulations presents challenges in amassing such voluminous datasets. Training a chatbot directly to design antennas for us is impractical. Thus, it becomes imperative to leverage certain techniques to facilitate the training of a valuable model using a smaller antenna dataset. In the generative workflow depicted in Fig.~\ref{classifier_in_generative}, these techniques encompass: 1) Converting the frequency responses obtained from EM simulations into scores based on design objectives. This conversion reshapes the training data \{images, scores\} into \{images, scores\}, inherently accommodating multi-objective design tasks while alleviating the precision demands on the classifier (surrogate model). This, in turn, trims down the dataset requirement. 2) Using the classifier to identify antenna geometry models with predicted scores surpassing a predetermined threshold value. Solely these chosen models undergo EM simulations. This strategy steers the median scores of the gathered dataset towards the target scores, effectively avoiding the accumulation of a substantial amount of ineffective data that deviates significantly from the performance objectives. 3) After completing a data collection batch, retraining the classifier involves the recalibration of not only model parameters and but also the threshold. Employing multiple iterations in this process enables the threshold value to approach the target score, thereby heightening the likelihood of obtaining optimal prototypes.

The combination of these techniques is aimed at leveraging a limited dataset to effectively discern whether geometric models warrant EM simulations, with the threshold for simulation worthiness gradually converging to the design target. Notably, the application of these techniques centers around the classifier. The generator in Fig.~\ref{classifier_in_generative} comprises a module responsible for converting geometrical models into corresponding 2D images. Complementing this is a discriminator, which plays the pivotal role of classifying images into either positive or negative categories. Only the geometrical model assigned a positive label are output from the generator. For the geometric models parameterized by the dimensions and positions of the fundamental components, as described in Appendix, factors such as overlapping and clipping makes the parameters of one geometric model non-unique. This intricacy prompts us to adopt an image-based representation rather than parameter arrays. Based on its huge success in computer vision~\cite{ lecun1995convolutional, simonyan2014very,he2016deep,krizhevsky2017imagenet,he2017mask}, CNN is chosen to construct the classifier.

\begin{figure*}[!t]
\centering \includegraphics[width=1 \columnwidth]{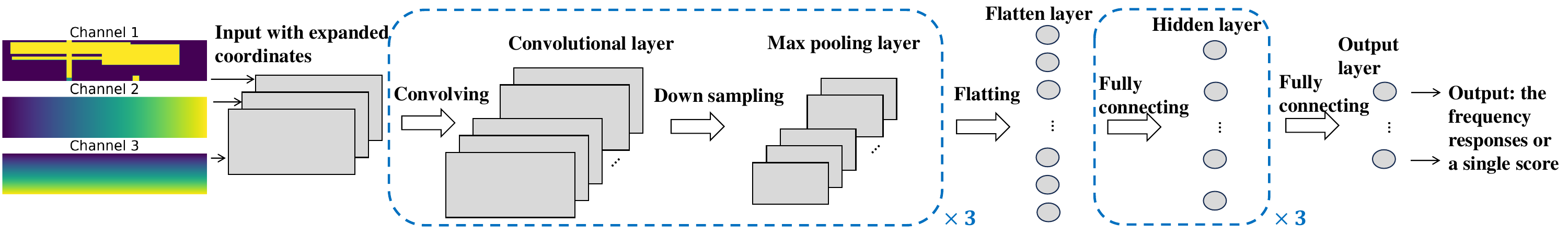}\\
\caption{
Structure of the CNN model. Source code can be find in the repository \url{https://github.com/yz564/antenna_ai}.
}
\label{cnn_structure}
\end{figure*}

Figure.~\ref{cnn_structure} illustrates the architecture of our CNN model, and the source code is accessible in the GitHub repository [\url{https://github.com/yz564/antenna_ai}] for reference.

The input comprises three channels although each geometric model entails only a single 2D image. The additional two channels represent the 2D expansions of the normalized x-coordinates and y-coordinates across the entire space \cite{AndrewSurrogate}. A visual representation of this input is shown in Fig.~\ref{cnn_structure}. Pixel values in these channels are constrained to the range of 0 to 1.  The first channel captures the geometric model of the antenna, utilizing values of 0 for the substrate, 1 for the metal, and 0.5 for the excitation port region. The second channel reflects the expanded x-coordinates, ranging from 0 on the left to 1 on the right. Similarly, the third channel encompasses the expanded y-coordinates, spanning from 0 at the top to 1 at the bottom. The normalized coordinates provide spatial information to our CNN, allowing the model to discern position-sensitive features within the image. 

The output consists of the frequency responses like the reflection coefficient S11, and a single score quantifying the deviation from the design target. These two outputs are distinct and operate independently in the CNN. During training, the optimization process results in a surrogate model by minimizing the frequency response loss, while it yields a classifier by minimizing the score loss. While it's possible to compute scores from predicted frequency responses, empirical testing suggests that directly training a score classifier offers enhanced accuracy. Table.~\ref{table:e0} compares them through an example involving 3500 training data and 500 new data for testing, with the classifier threshold set as the median of the training data. During model training, the focal point is on minimizing the mean squared error (MSE) loss. Meanwhile, the CNN classifier's performance is gauged using the true positive (TP) rate and false positive (FP) rate.

The training of this CNN model incorporates dropout and weight decay regularization techniques, strategically implemented to counter overfitting. A learning rate scheduler is employed, which dynamically adjusts the learning rate when the validation loss reaches a plateau. Additionally, hyperparameters like the number of neurons, batch size, are fine-tuned by relying on the TP rate and FP rate, which serve as our primary metric for evaluating the performance of the classifier.

\begin{table}[h!]
\caption{Comparison of CNN Models with different output}
\label{table:e0}
\centering
\begin{tabular}{|c|c|c|c|c|c|c|c|} 
\hline
CNN model & \multicolumn{3}{c|}{Train} & \multicolumn{3}{c|}{Test} \\
\cline{2-7}
& \multicolumn{1}{c|}{MSE loss} & TP rate & FP rate & \multicolumn{1}{c|}{MSE loss} & TP rate & FP rate \\
\hline
S11 at target frequencies & 1.42 & 0.85 & 0.43 & 3.74 & 0.97 & 0.84 \\
\hline
Full S11 & 0.93 & 0.81 & 0.30 & 1.78 & 0.95 & 0.69 \\
\hline
Score & 0.11 & 0.86 & 0.17 & 0.19 & 0.94 & 0.58 \\
\hline
\end{tabular}

 \end{table}

\section{Selection of Promising Dimensions}
\label{dimension}

In wireless consumer electronics, many antenna designs can be broken down into fundamental geometric components, often resembling fixed-width and height rectangles. For instance, consider a loop-shaped PCB antenna operating within the frequency range of 2.4-2.5 GHz and 5.1-5.9 GHz. This antenna can be deconstructed into five individual rectangles, each defined by specific dimensions listed in Set 1 of Table~\ref{table:e1}. Before arranging these fixed-dimension components within a 30 mm $\times$ 6 mm area to meet the new performance objective of achieving $|S_{11}|<-6$ dB at 2.4-2.5 GHz and 5.1-7 GHz, it is essential to evaluate the suitability of these dimensions for the revised design criteria. Therefore, we have introduced an additional four sets of dimension variations, as outlined in Table~\ref{table:e1}, to compare their potential of achieving the design goal.

\begin{table}[h!]
\caption{Candidate Dimension Set for Example 1}
\label{table:e1}
\centering
\begin{tabular}{|c | c |  c | c | c | c | c |} 
 \hline
Dimension & \multicolumn{5}{c|}{Set [i]} \\ 
\cline{2-6}
(width, height) & 1 & 2 & 3 & 4 & 5 \\
 \hline
 Component 1 & (0.75,5.49) &  (1,5) & (1,4) & (2,5) & (1,4) \\
 \hline
 Component 2 & (16.87,1.7) &  (12,2) & (6,4) & (16,3) & (18,2) \\
 \hline
 Component 3 & (11.38,3) &  (16,2) & (8,3) & (12,2) & (5,3) \\
 \hline
 Component 4 & (18.63,0.56) &  (15,1) & (10,2) & (7,2) & (9,1) \\
 \hline
 Component 5 & (0.99,2.43) &  (2,4) & (2,3) & (2,5) & (1,4) \\
 \hline
 \end{tabular}
 \end{table}

\begin{figure}[!t]
\centering \includegraphics[width=0.8 \columnwidth]{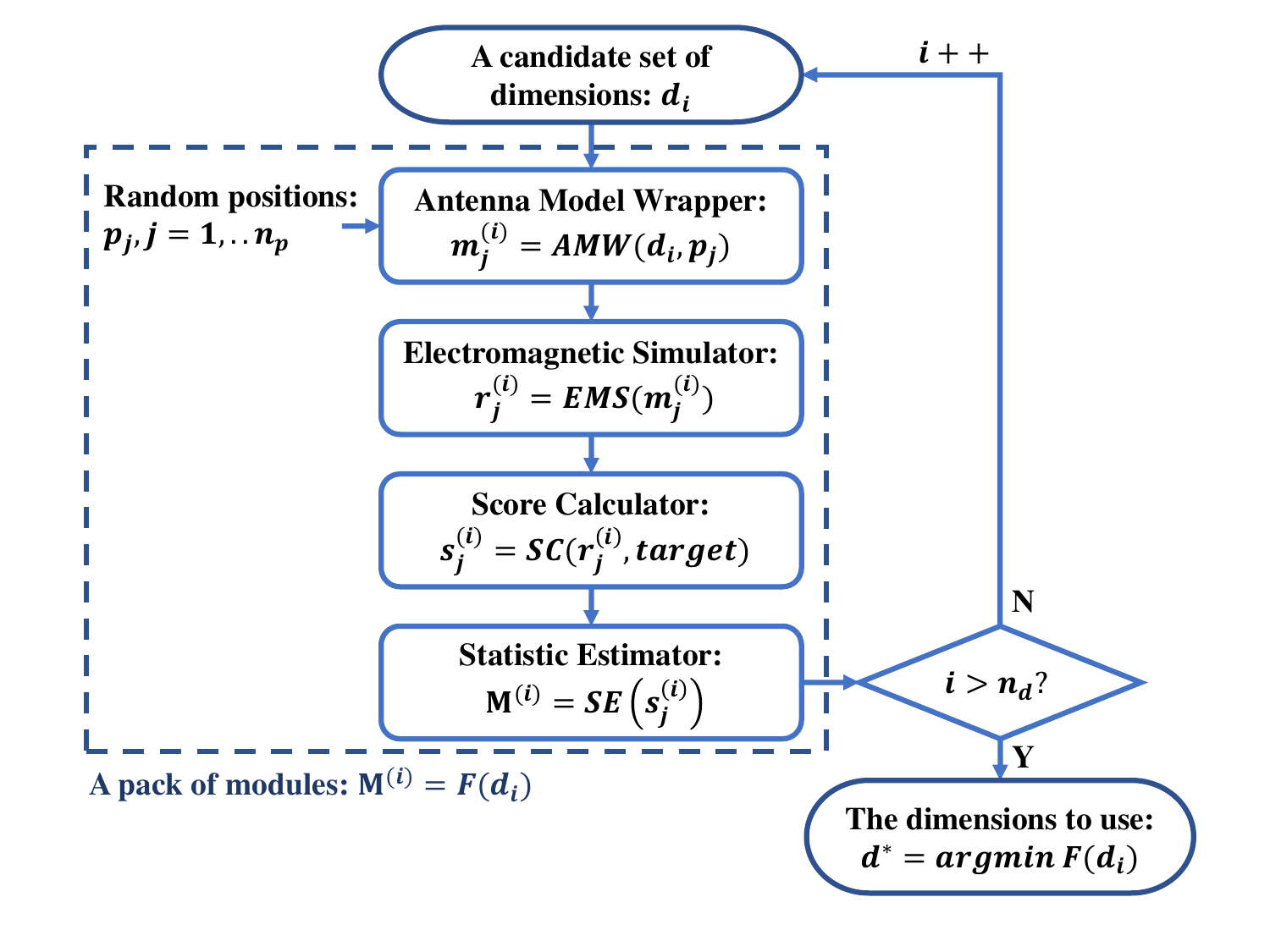}\\
  \caption{Workflow of the dimension selector. }\label{dimension_selector}
\end{figure}

Our metric is the median score of 100 samples, wherein fixed-dimension components are randomly positioned. As illustrated in Fig.~\ref{dimension_selector}, a pack of modules estimates a statistical median score $M^{(i)}$ for each candidate set of dimensions $d_i$. The first module is an antenna model wrapper that takes the dimensions and positions as input and outputs an antenna model within the simulation environment. The second module is an electromagnetic simulator to produce the result $r_j^{(i)}$. This result is then processed by the third module, which computes a score reflecting the maximum deviation from the target values within the designated frequency range. The last module is a statistic estimator, determining the median score from the gathered data. For each distinct set of fixed-dimension components, we conduct simulations for $n_p=100$ antenna models, each with randomized component positions. The dimension set associated with the smallest median score is ultimately selected for further consideration.

\begin{figure}[!t]
\centering
\subfloat[]{\includegraphics[width=0.45\columnwidth]{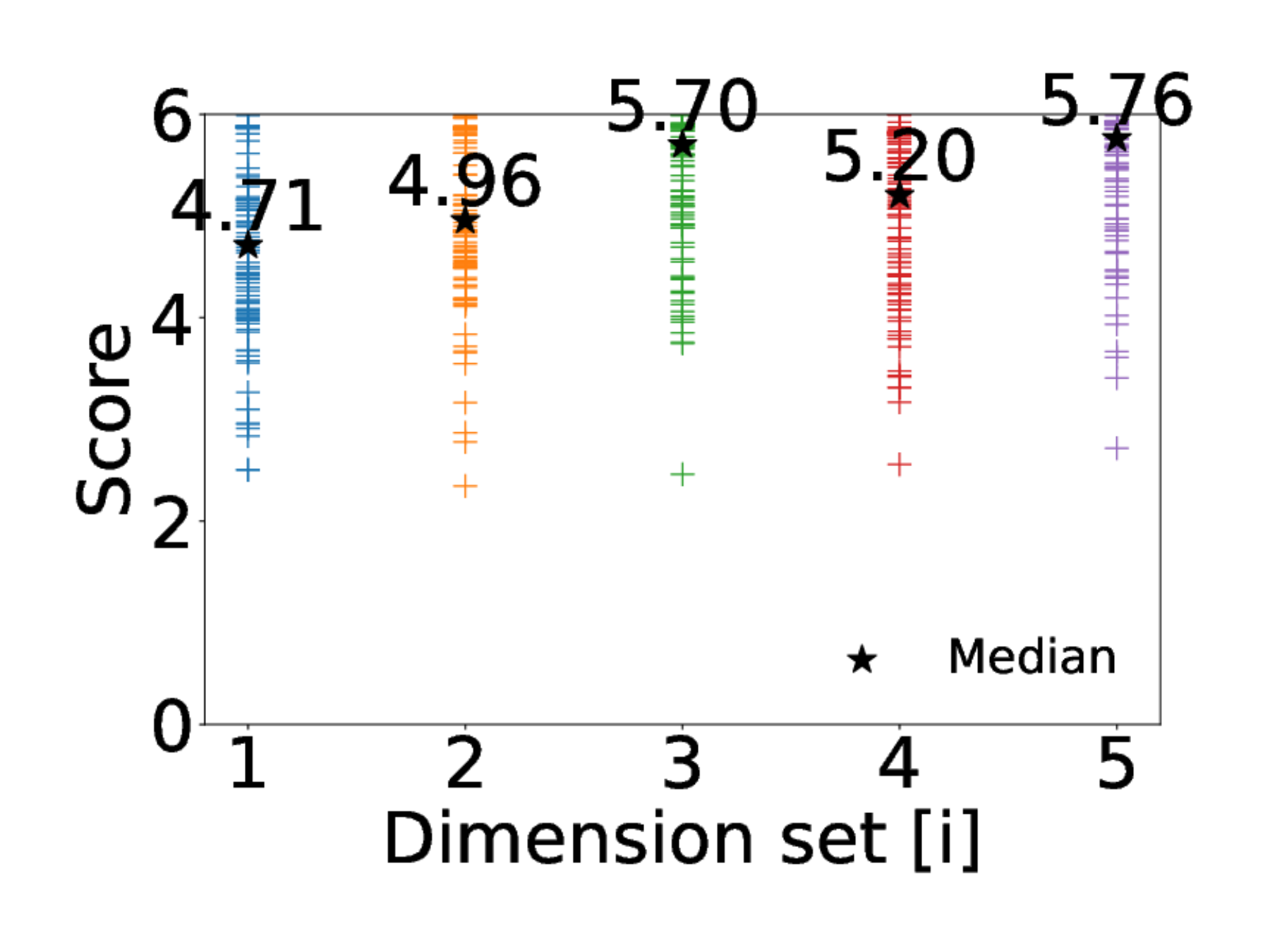}
\label{ex1_dim_points}}
\subfloat[]{\includegraphics[width=0.45\columnwidth]{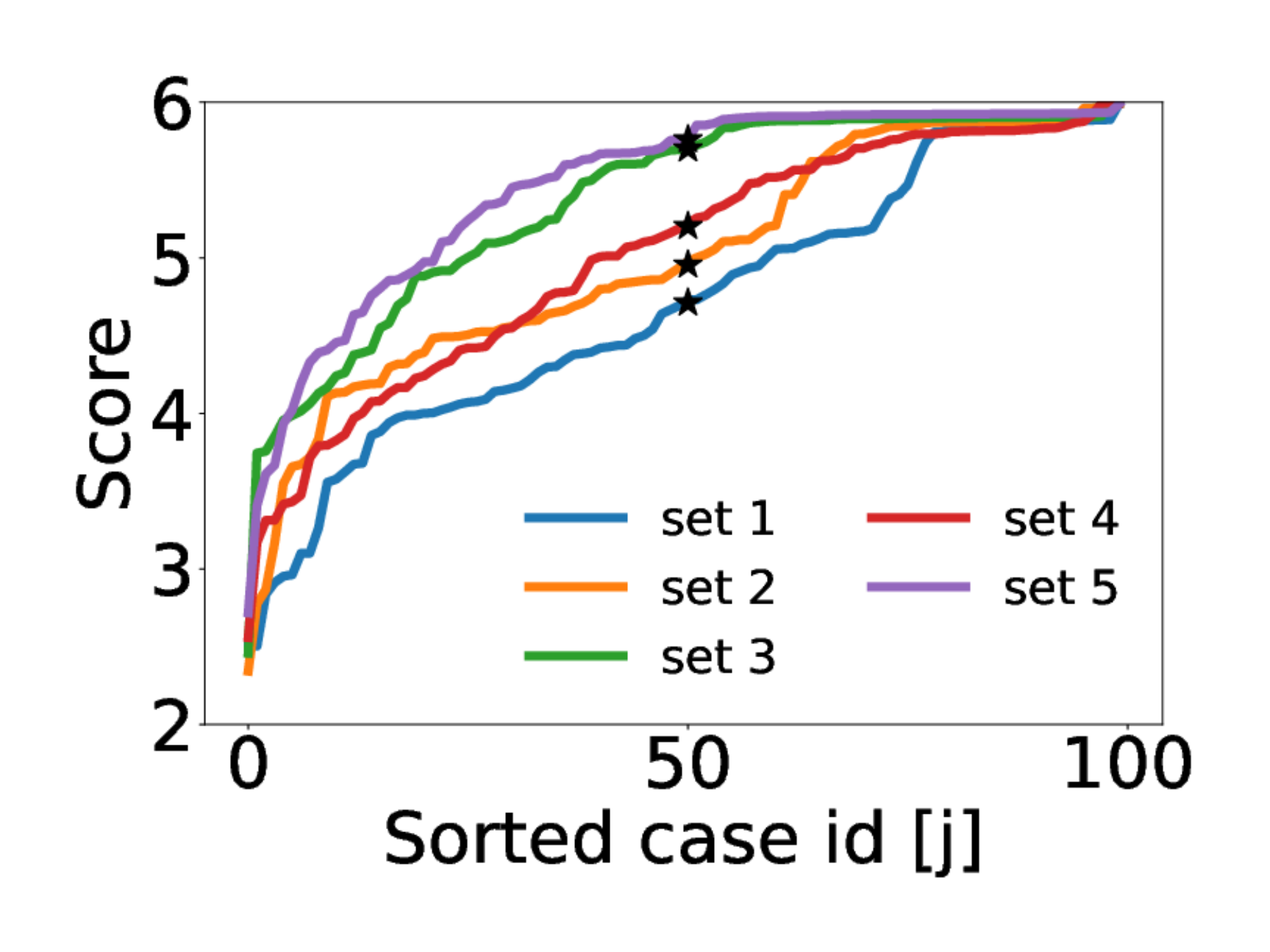}
\label{ex1_dim_curves}}
\caption{Comparison of the sample score distribution for five sets of dimensions. (a) Each '+' sign represents one simulation data. (b) Each curve represents 100 random position simulation data with the fixed dimensions.}
\label{ex1_dim}
\end{figure}

Following this procedure, we present the comparison of the five candidate sets in Fig.~\ref{ex1_dim_points} and Fig.~\ref{ex1_dim_curves}, offering different viewpoints of the same dataset. In Fig.~\ref{ex1_dim_points}, the discrete horizontal axis represents the candidate IDs, indicating that the data points on a vertical line have the same component dimensions. The diverse scores are associated with the positions of these components. The median score of the first set is the smallest. In Fig.~\ref{ex1_dim_curves}, there are five curves, each corresponding to one of the five candidate dimension sets. These curves are monotonic because they are ordered according to the respective scores. Considering the distribution of samples, the first set of dimensions should also be selected from the five choices ($d^*=d_1$).

\section{Placement of Selected Dimension Components}
\label{position}

\begin{figure}[!t]
\centering \includegraphics[width=0.8 \columnwidth]{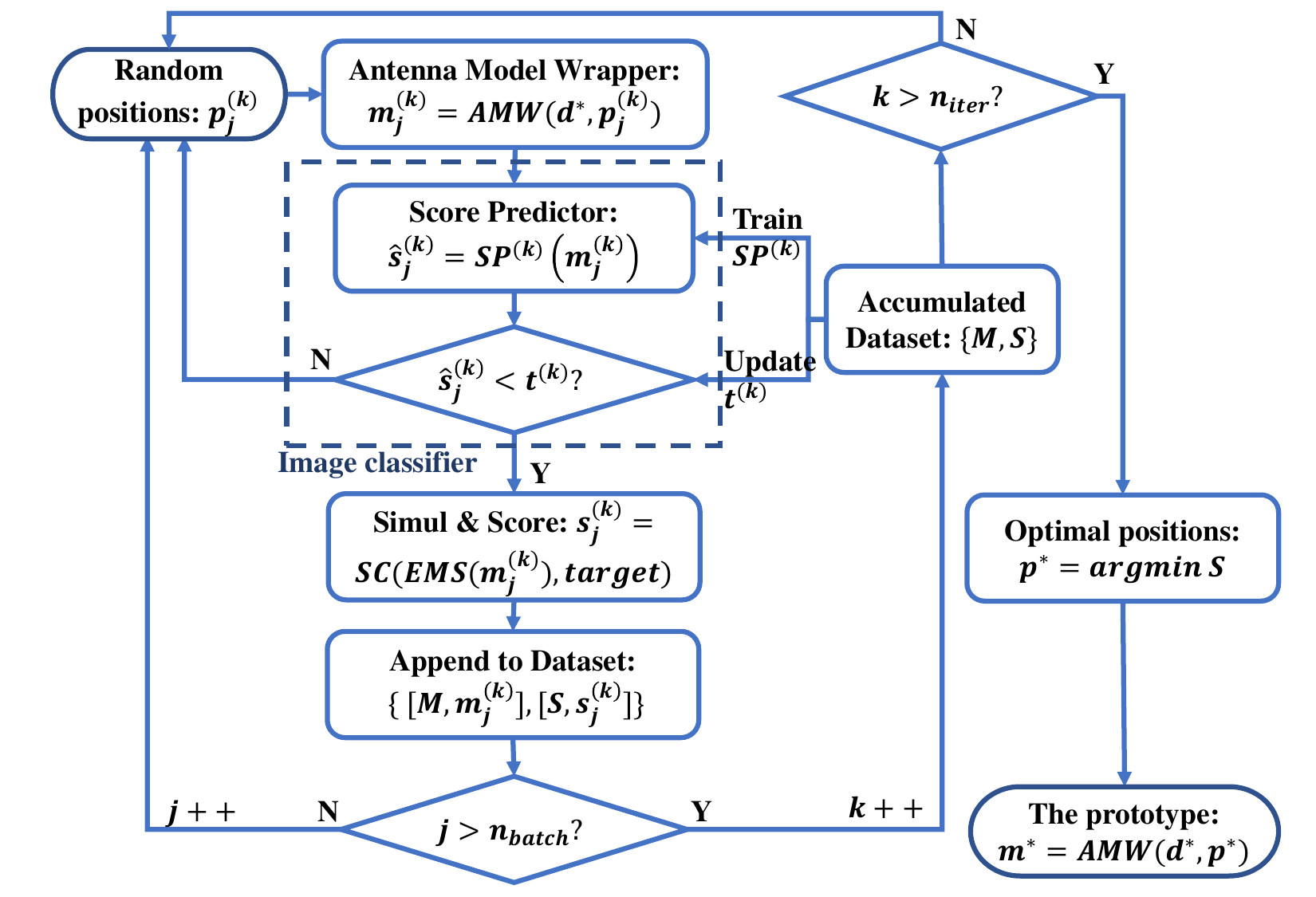}\\
  \caption{Workflow of the antenna model generator using components with the fixed dimensions. }\label{position_generator}
\end{figure}

Empowered by the dimension selector, we are now more confident in using these fixed-dimension component for prototype development. The next step is to place these components within confined spaces, following the method illustrated in Fig.~\ref{position_generator}. The antenna model wrapper takes the chosen dimensions ($d^*$) and random positions ($p_j^{(k)}$), creating $m_j^{(k)}$. Instead of directly interfacing with the simulator, this model output is first channeled to a classifier for preliminary evaluation. The classifier comprises two parts: a CNN as depicted in Fig.~\ref{cnn_structure}, designed to predict a score based on the 2D antenna image; and a decision-making module with a designated threshold. The score predictor (denoted as {SP}$^{(k)}$) is trained using accumulated simulation data and the threshold value ($t^{(k)}$) can also be derived from the dataset. The superscript $k$ indicates the iterative stages of the classifier. The classifier updates are triggered after collecting $n_{batch}$ simulation data points. Only antenna models that successfully pass through the filter are subjected to simulation. This selective simulation process increases the likelihood of generating antenna models that surpass the threshold compared to randomly generated models without the filtering mechanism, as long as the trained CNN remains effective. After $n_{iter}$ iterations, i.e. $n_{batch} \times n_{iter}$ simulations, we can use one of the top-performing simulated antenna models to serve as the prototype.

\begin{figure}[!t]
\centering
\subfloat[]{\includegraphics[width=0.45\columnwidth]{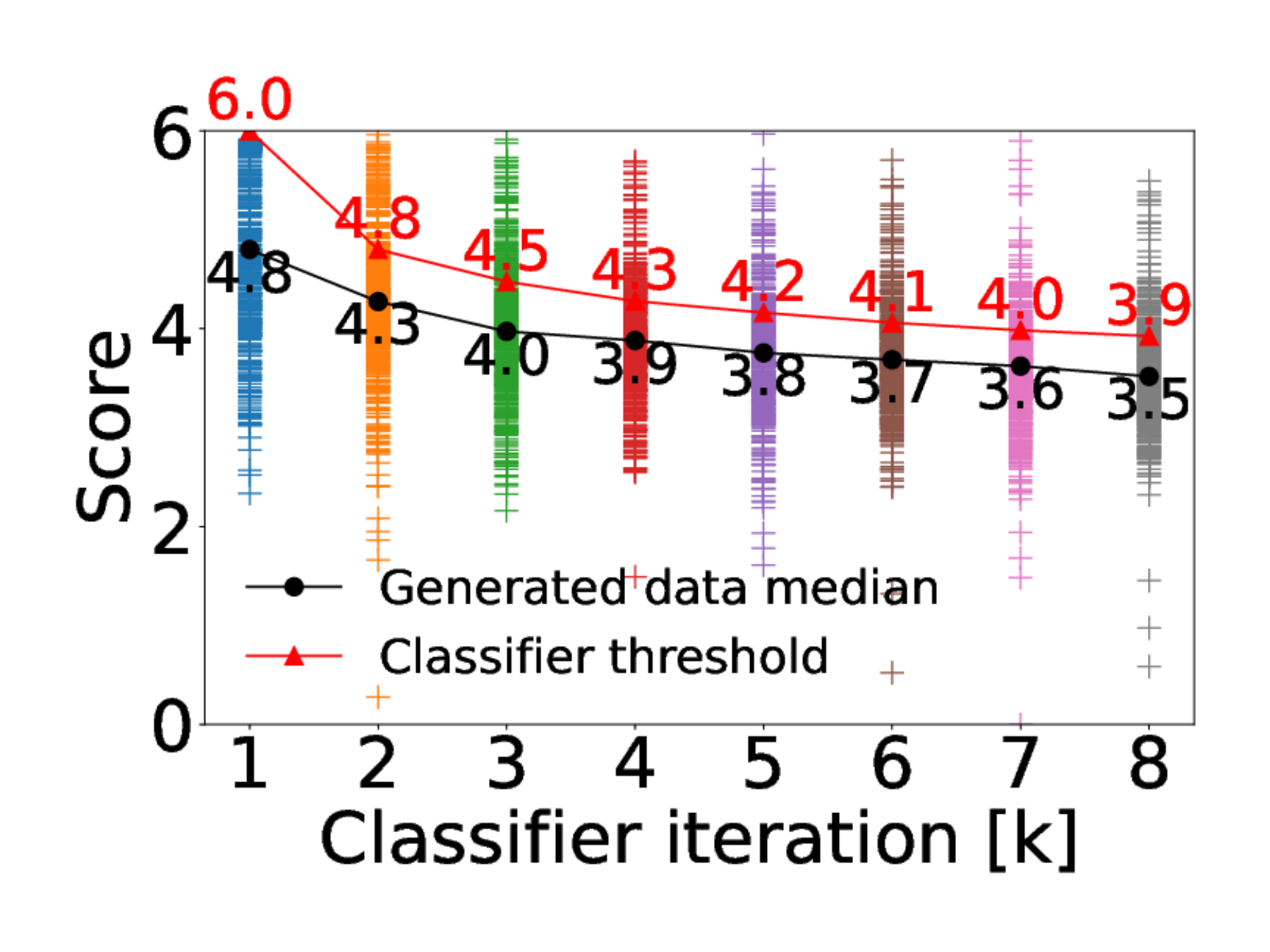}
\label{ex1_pos_points}}
\subfloat[]{\includegraphics[width=0.45\columnwidth]{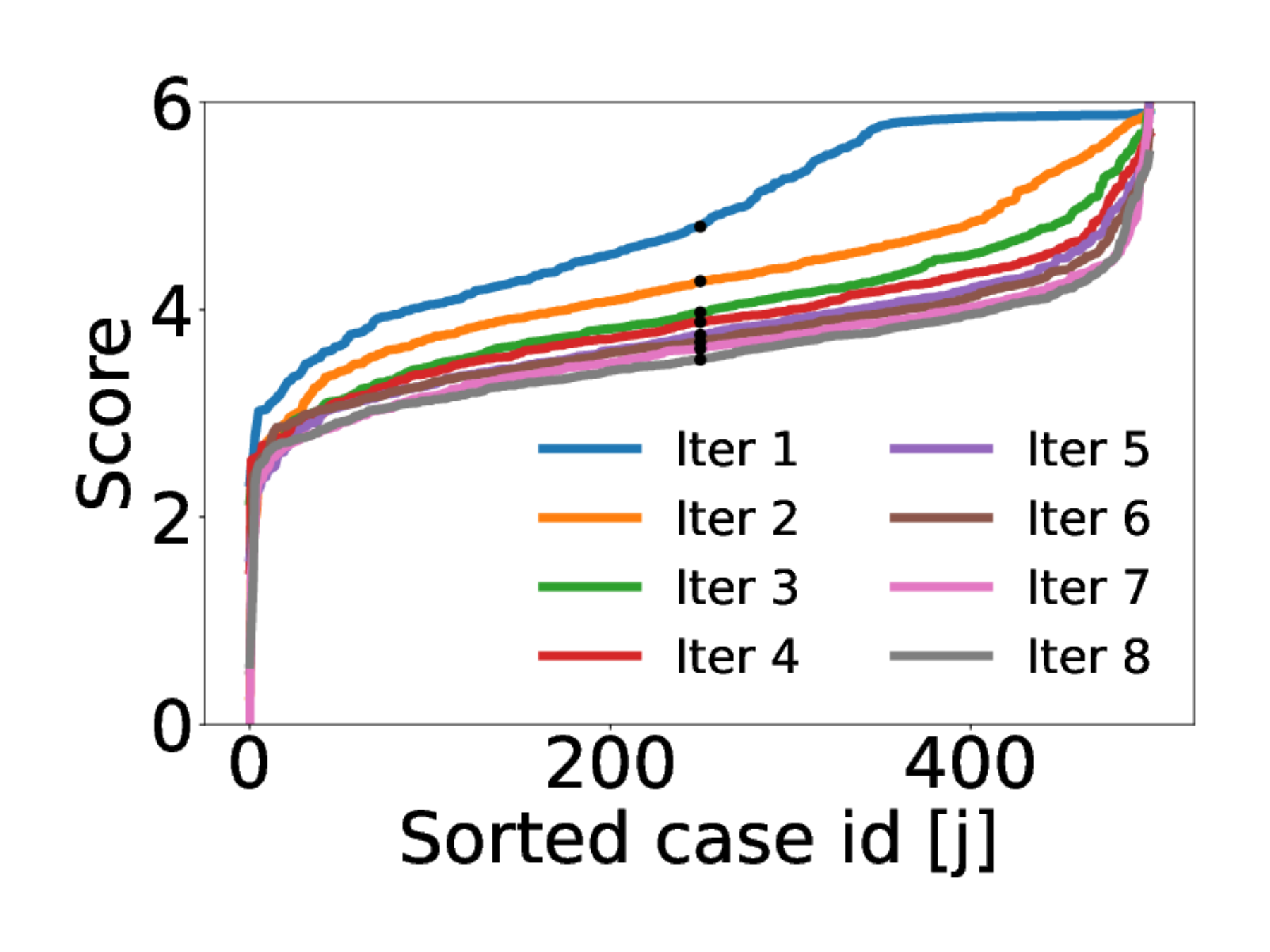}
\label{ex1_pos_curves}}
\caption{Score distribution comparison of the generated data over eight iterations. (a) Each '+' sign represents one simulation data. (b) Each curve represents 500 filtered position simulation data with the selected dimensions.}
\label{ex1_pos}
\end{figure}

\begin{figure}[!t]
\centering
\subfloat[]{\includegraphics[width=0.45\columnwidth]{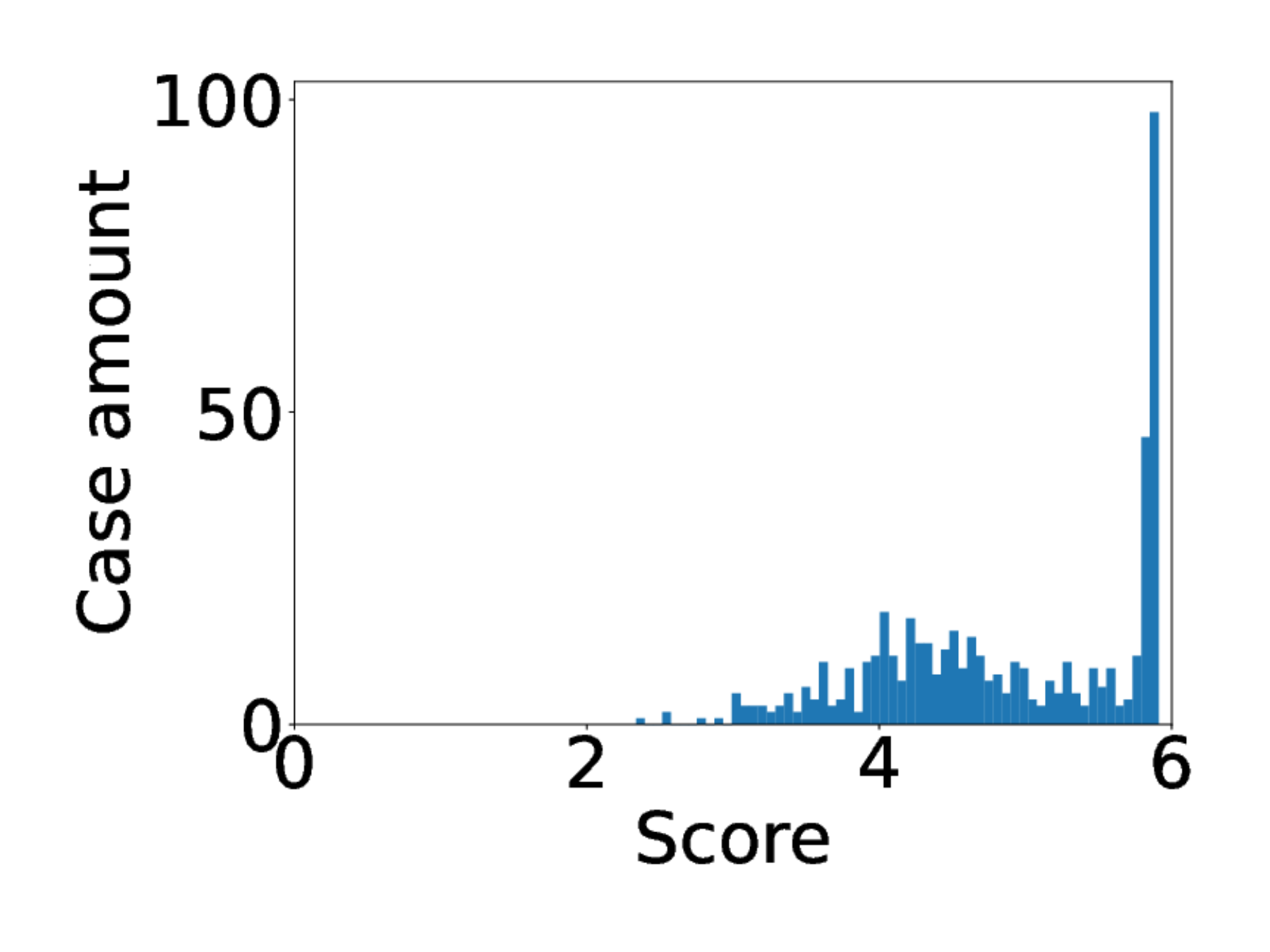}
\label{ex1_hist_first}}
\subfloat[]{\includegraphics[width=0.45\columnwidth]{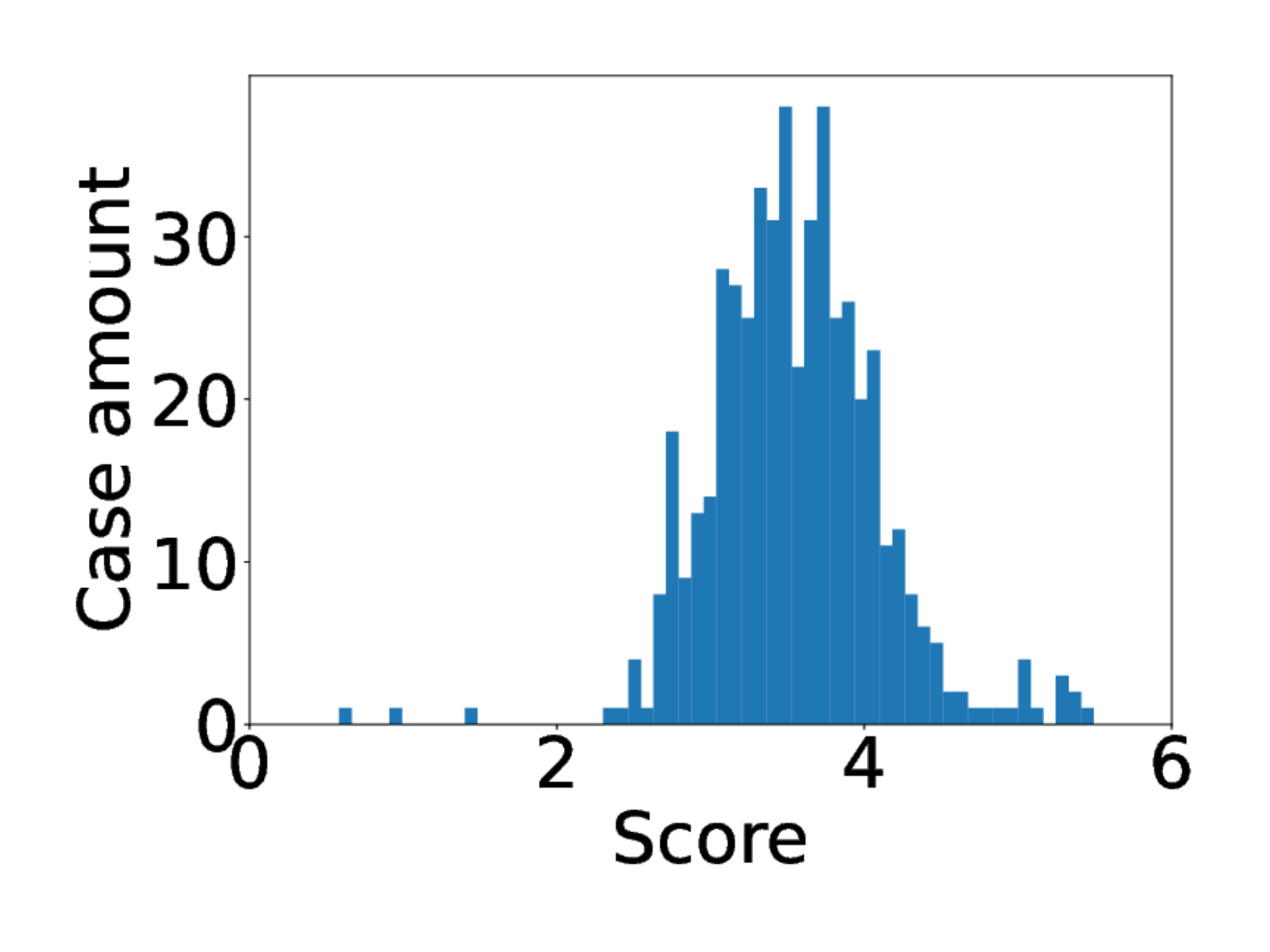}
\label{ex1_hist_last}}
\caption{Comparison of the score histograms. (a) The initial 500 purely random simulation data. (b) The last 500 simulation data filtered in the eighth iteration.}
\label{ex1_hist}
\end{figure}

Fig.~\ref{ex1_pos_points} and Fig.~\ref{ex1_pos_curves} shows the distribution of scores across each iteration. The threshold value is selected to be the median score of the accumulated dataset, except the initial iteration $t^{(1)}$=6 where no data is available to train the filter. The median score of the newly simulated data consistently surpasses the threshold value, indicating the effective operation of the filter. Furthermore, examining the histograms, as depicted in Fig.~\ref{ex1_hist}, reveals that the distribution of antenna models produced by the generator has shifted closer to the design target.

\begin{figure}[!t]
\centering
\subfloat[]{\includegraphics[width=0.5\columnwidth]{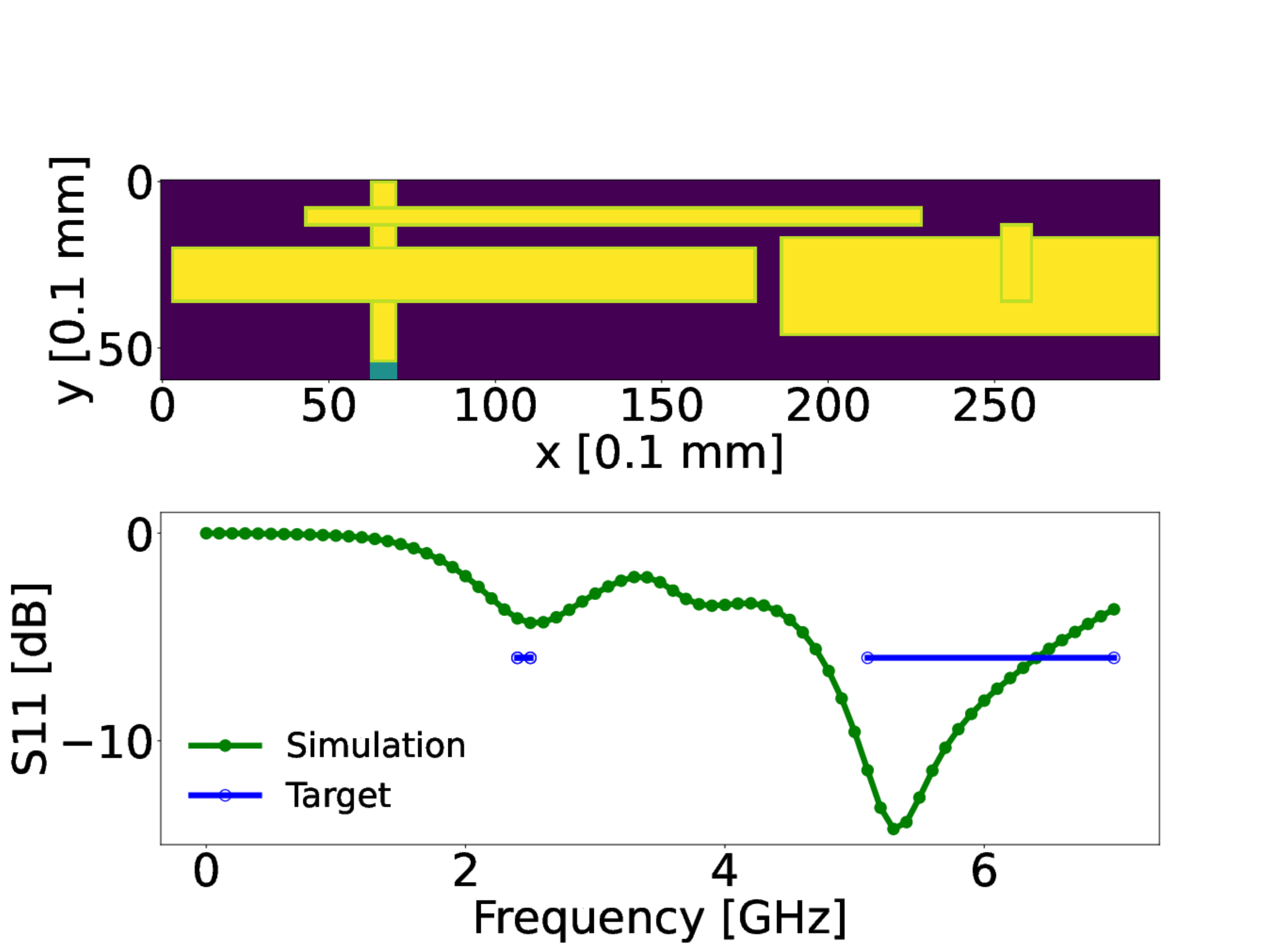}
\label{ex1_280}}
\subfloat[]{\includegraphics[width=0.5\columnwidth]{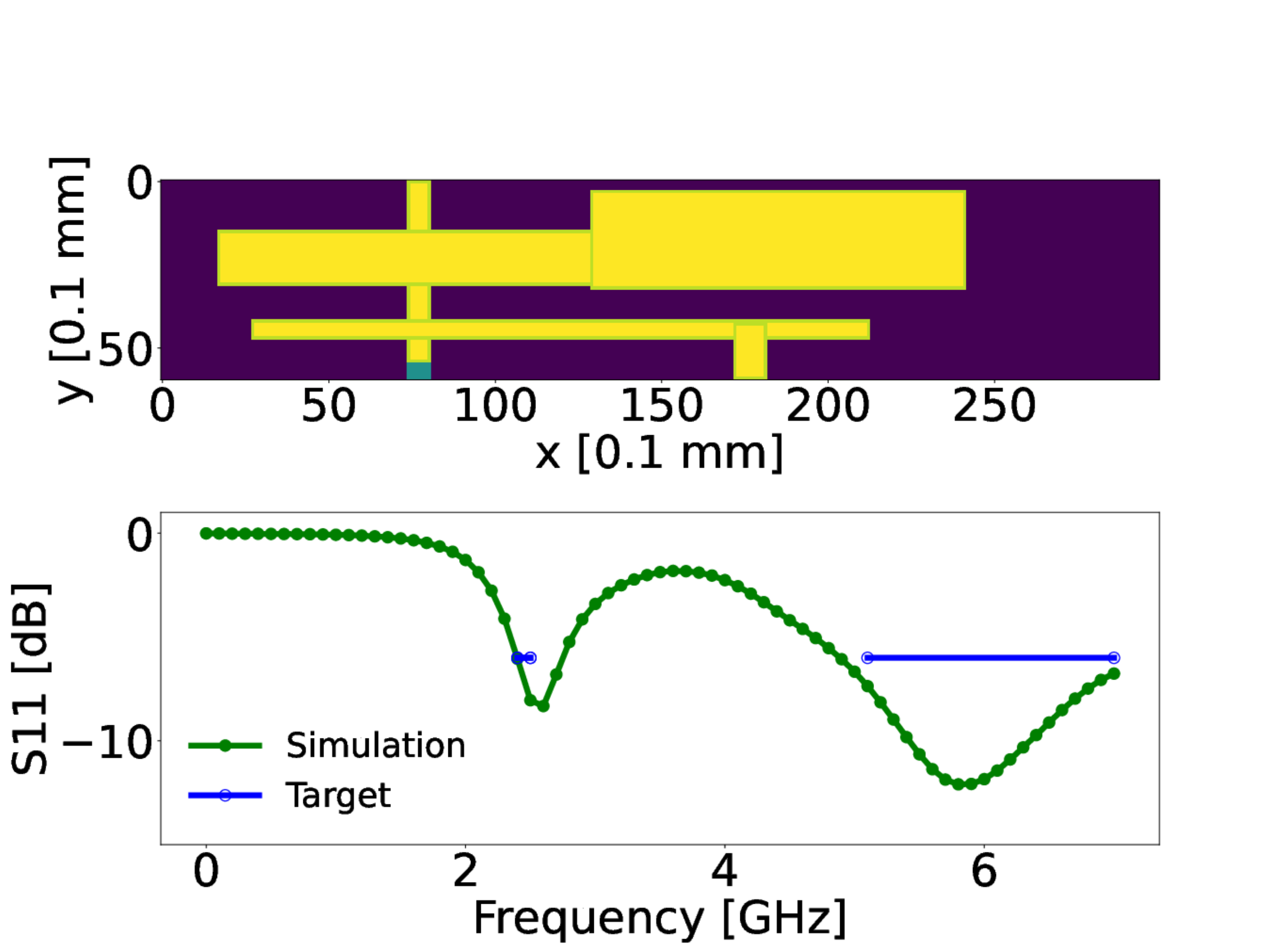}
\label{ex1_3411}}
\caption{The geometric model image and the corresponding simulation result. (a) The best of the initial 500 random data. (b) The best of the 4000 accumulated data.}
\label{ex1_results}
\end{figure}

\begin{figure}[!t]
\centering \includegraphics[width=0.7 \columnwidth]{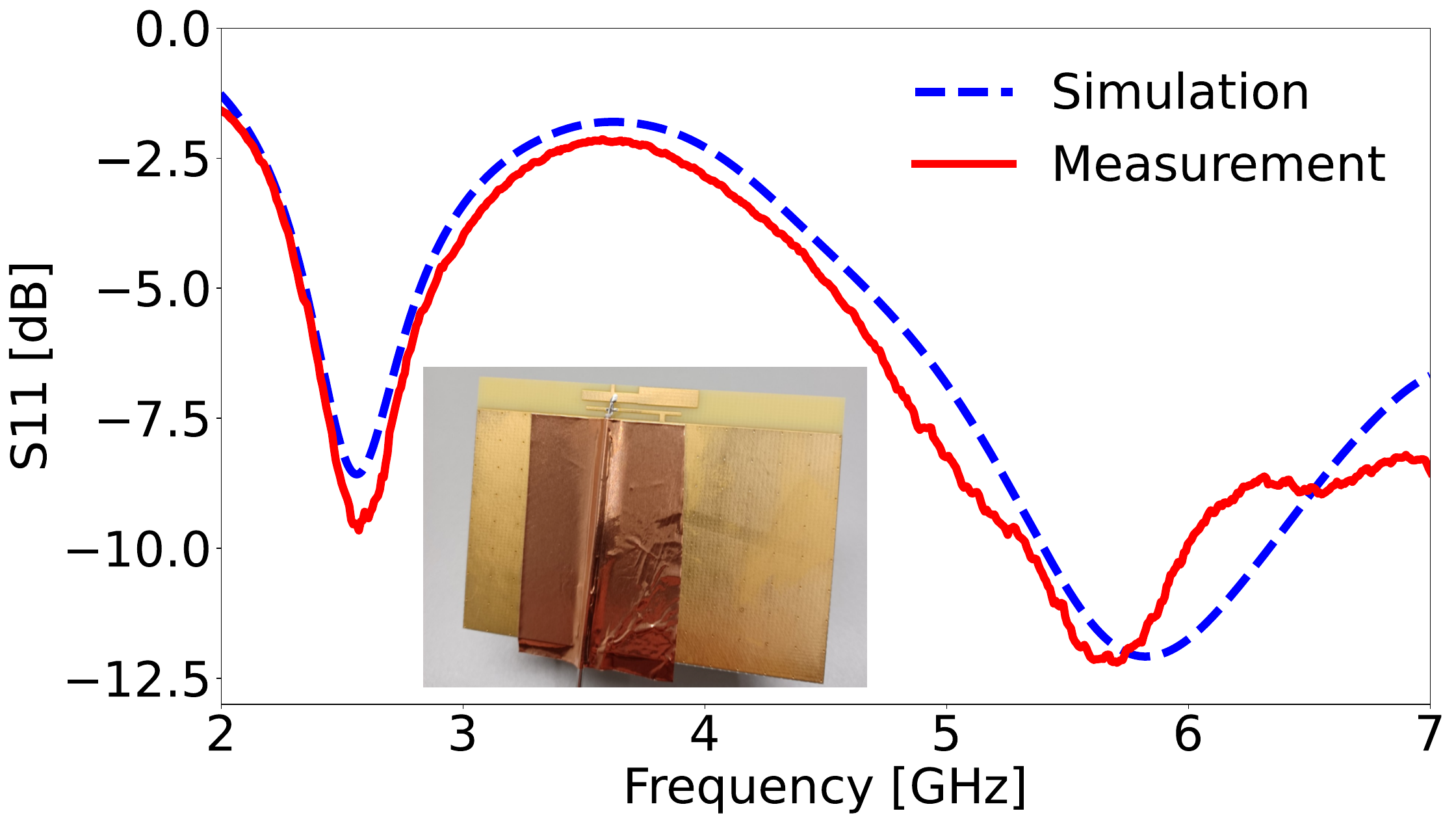}\\
  \caption{Comparison of the simulation and measurement for the first example. }\label{example1_pcb}
\end{figure}

Fortunately, one case in the seventh iteration already meets the target, as shown in Fig.~\ref{ex1_3411}. This prototype waives the last optimization in Fig.~\ref{overall}. For comparison, Fig.~\ref{ex1_280} shows the best case in the initial 500 simulations. The PCB measurement aligns well with the simulation, as shown in Fig.~\ref{example1_pcb}.

\section{Antenna Example for AR Glasses}
\label{real_example}

\begin{figure}[!t]
\centering \includegraphics[width=0.9 \columnwidth]{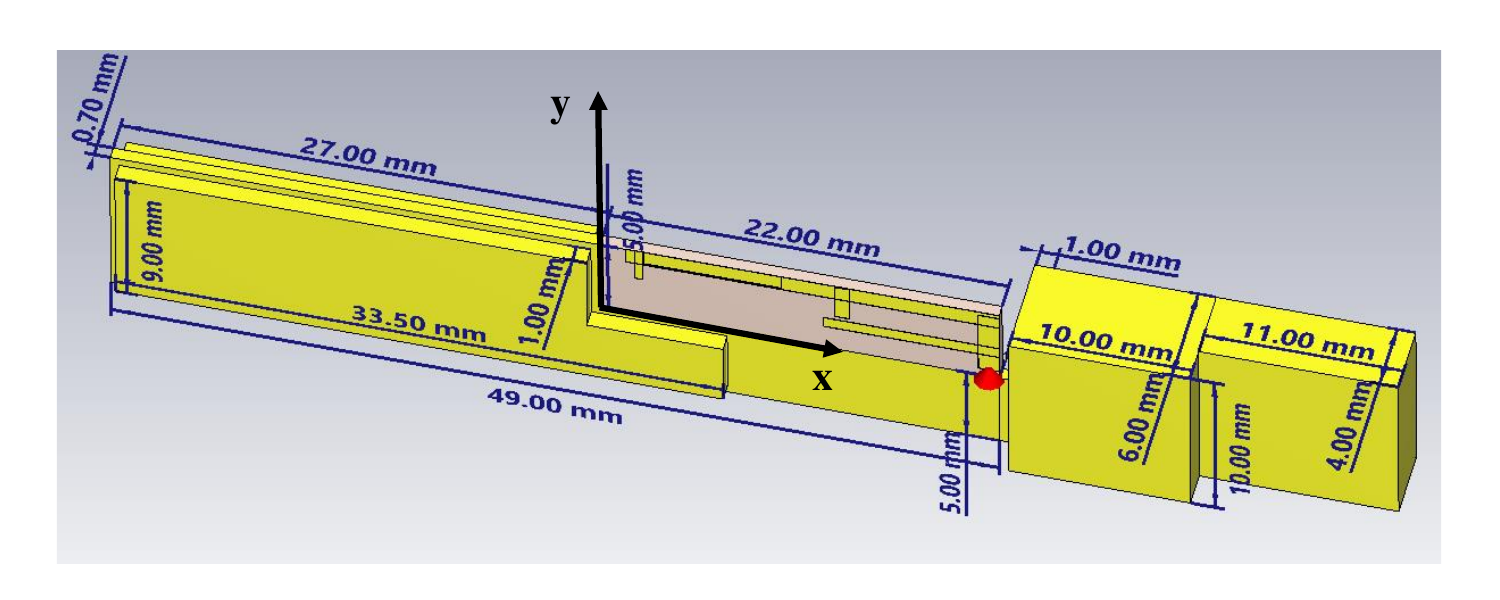}\\
  \caption{A more realistic example with compacter design spaces and a more complicated environment. }\label{example2}
\end{figure}

For some Metaverse products like AR glasses, the antenna volume requirement is even smaller than the previous example, for example 22 mm $\times$ 5 mm, as shown in Fig.~\ref{example2}. The surrounding environment becomes more complicated too, including a few metal boxes to emulate RF shielding cans and other components. The remaining design targets are the same in order to support traditional dual-band WiFi as well as the faster speed WiFi 6E. In this second example, we initially employ the same two-step approach as before to propose an antenna prototype. Subsequently, we leverage trust region optimization to finely adjust the geometric parameters of the prototype.

\subsection{Prototype}
\label{example_prototype}

\begin{figure}[!t]
\centering
\subfloat[]{\includegraphics[width=0.45\columnwidth]{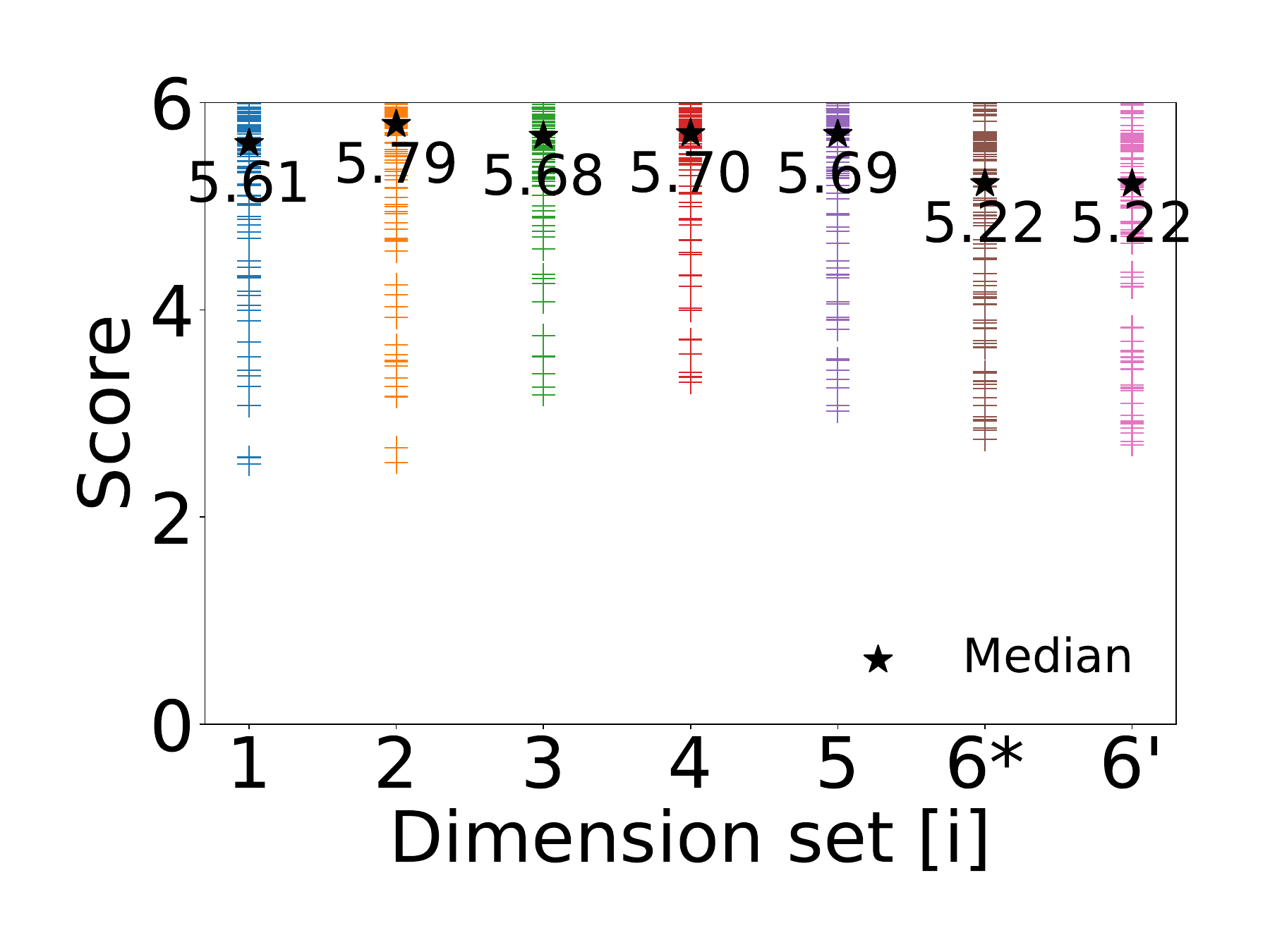}
\label{ex2_dim_points}}
\subfloat[]{\includegraphics[width=0.45\columnwidth]{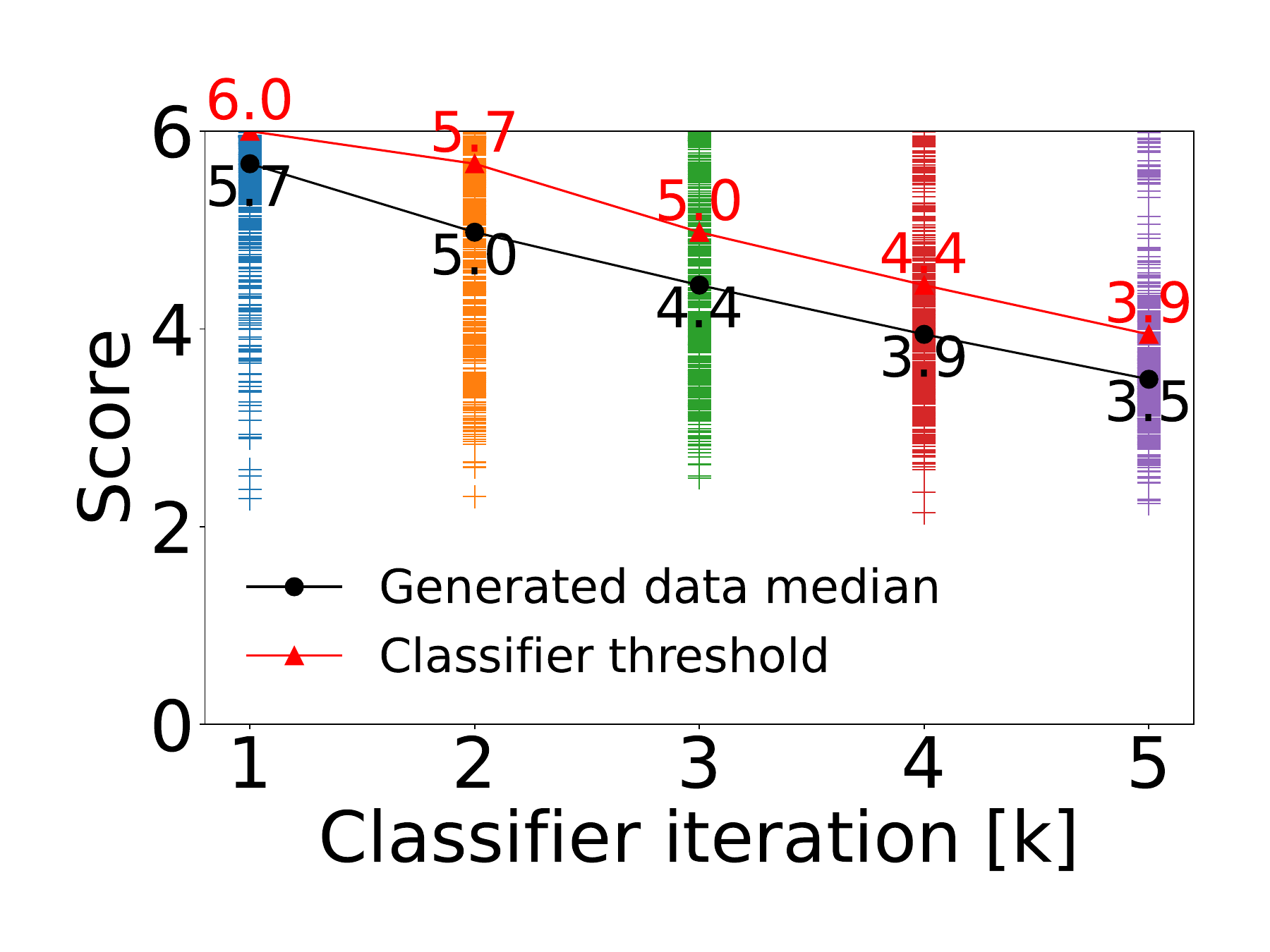}
\label{ex2_pos_points}}
\caption{Distribution of the simulated data in the two steps. (a) The random-position sample data from the dimension selector. (b) The filtered-position generated data from the antenna model generator.}
\label{ex2_dim}
\end{figure}

In this example, we use six rectangles as fundamental components. The dimensions of these rectangles are presented in Table~\ref{table:e2} across five distinct sets, denoted as set $i$ = 1 to 5. Set 6 will be discussed in Subsection \ref{analysis} and is temporarily excluded from consideration. Figure.~\ref{ex2_dim_points} shows the distribution of the random simulation data from the dimension selector. With the selected dimension set 1, Fig.~\ref{ex2_pos_points} displays the generated data after fifth iterations of the classifiers. In this instance, the threshold value is updated using the median score from the previous 500 generated data.

\begin{table}[h!]
\caption{Candidates of Component Dimensions for Example 2}
\label{table:e2}
\centering
\begin{tabular}{|c | c |  c | c | c | c | c | c| c|} 
 \hline
Dims & \multicolumn{6}{c|}{Set [i]} \\ 
\cline{2-7}
(w, h) & 1 & 2 & 3 & 4 & 5  & 6' \\
 \hline
 C1 &  (0.75,4) &  (1,4) & (1,4) & (1,5) & (1,5) &  (1,4) \\
 \hline
 C2 &  (17.64,1.7) &  (15,2) & (16,1) & (10,2) & (9,1) &  (21,2) \\
 \hline
 C3 &  (11.38,3) &  (8,3) & (5,4) & (12,3) & (10,1) &  (8,3) \\
 \hline
 C4 &  (18,0.56) &  (12,1) & (10,2) & (15,1) & (8,3) &  (18,1) \\
 \hline
 C5 &  (0.99,2.43) &  (2,3) & (1,4) & (3,4) & (7,2) &  (1,2) \\
 \hline
 C6 &  (0.6,4) &  (1,4) & (3,1) & (1,2) & (1,3) &  (1,3) \\
 \hline
 \end{tabular}
 \end{table}

\begin{figure}[!t]
\centering
\subfloat[]{\includegraphics[width=0.5\columnwidth]{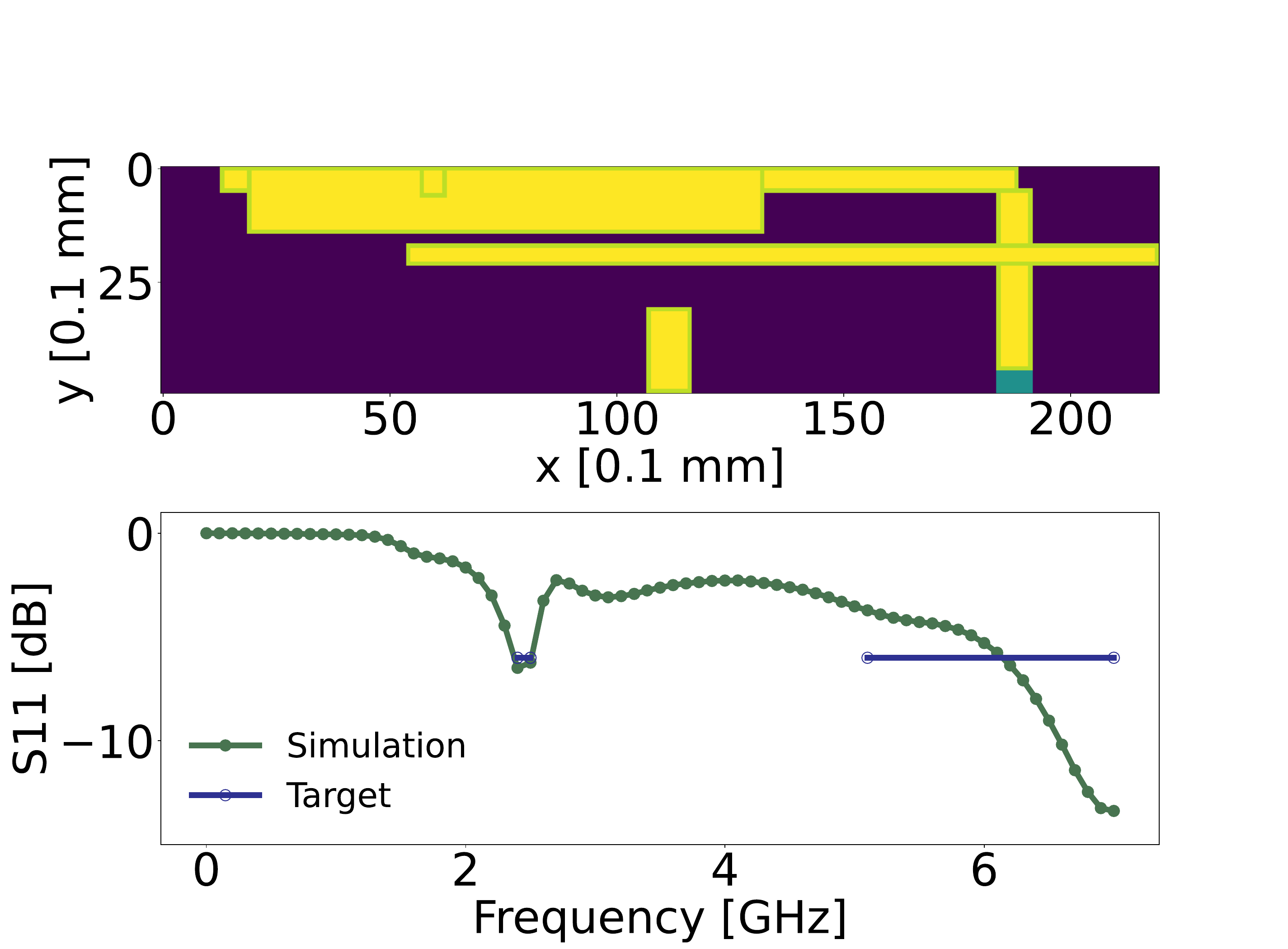}
\label{ex2_331}}
\subfloat[]{\includegraphics[width=0.5\columnwidth]{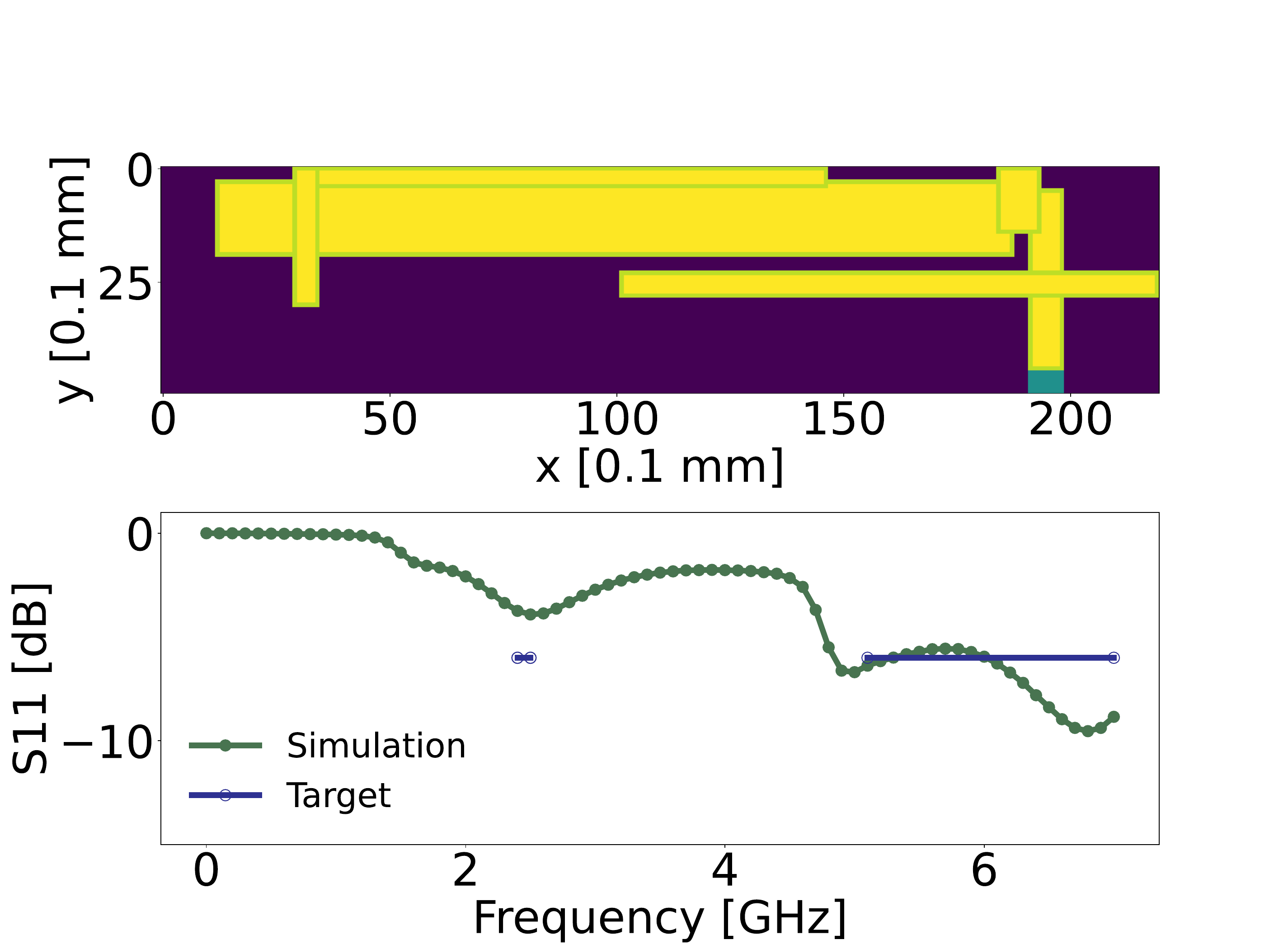}
\label{ex2_2173}}
\caption{Two top-performing geometric models in the 2500 simulation data. (a) One meets the higher frequency band requirement. (b) Another almost meets the higher frequency band requirement.}
\label{ex2_results}
\end{figure}

While none of the simulated antenna models reach the intended design target, there are several cases with scores falling within the 2.2 to 2.3 range. Specifically, in Fig.~\ref{ex2_331}, we observe a generated case that successfully meets the goal in the lower frequency band. Additionally, Fig.~\ref{ex2_2173} presents another generated case that comes close to meeting the goal in the higher frequency band. For the subsequent optimization efforts, we have chosen the latter case as the prototype.

\subsection{Optimization}
\label{example_optimization}

\begin{figure}[!t]
\centering \includegraphics[width=0.7 \columnwidth]{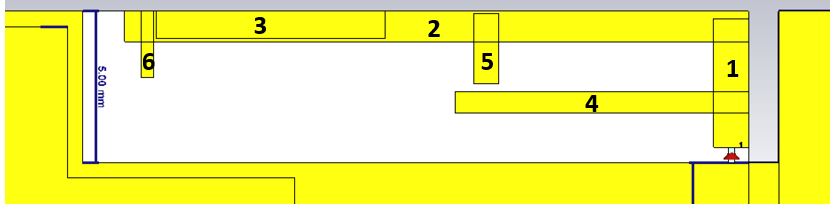}\\
  \caption{The final antenna model after optimization. }\label{ex2_opt}
\end{figure}

\begin{figure}[!t]
\centering \includegraphics[width=0.8 \columnwidth]{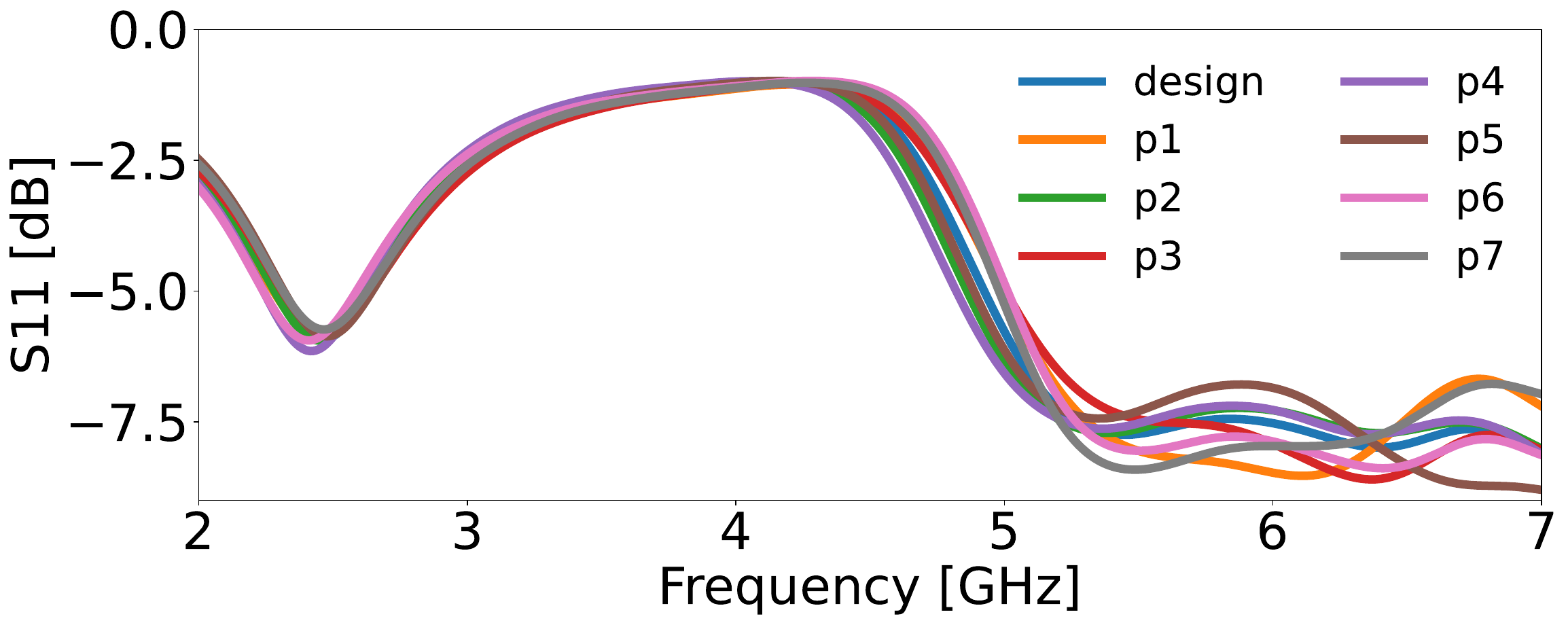}\\
  \caption{Tolerance study: simulation S11 curves for the antenna model geometric parameters with random perturbation within 10\%. }\label{ex2_tol}
\end{figure}

After an antenna prototype or the type of the antenna design is determined, traditional optimizers can be applied for further performance improvement. For the prototype as shown in Fig.~\ref{ex2_2173}, we use the trust region optimizer to tune the geometric parameters. Within 700 simulations, an antenna model that meets the target has been found, as depicted in Fig.~\ref{ex2_opt}. Table~\ref{table:e2_opt} lists dimensions and positions of the six components. This design has sufficient tolerance to the perturbation as illustrated in Fig.~\ref{ex2_tol}, suggesting the final design is reliable.

 \begin{table}[h!]
\caption{Geometric Parameters of the Final Simulation Model}
\label{table:e2_opt}
\centering
\begin{tabular}{|c |c | c | c | c | c | c | c | } 
 \hline
\multicolumn{2}{|c|}{Parameters} & C1 & C2 & C3 & C4 & C5 & C6  \\ 
\hline
{Dimensions}& w & 1.2 & 21.1 & 7.6 & 17.8 & 0.8 & 0.4 \\
\cline{2-8}
& h & 4.2 & 2.2 & 3.2 & 0.7 & 2.3 & 2.9 \\
\hline
{Positions} & x & 20.8 & 1.4 & 2.4 & 12.3 & 12.9 & 1.9 \\
\cline{2-8}
& y & 0.5 & 4.0 & 4.1 & 1.6 & 2.6 & 2.8 \\
\hline
 \end{tabular}
 \end{table}

\begin{figure}[!t]
\centering
\subfloat[]{\includegraphics[width=0.8\columnwidth]{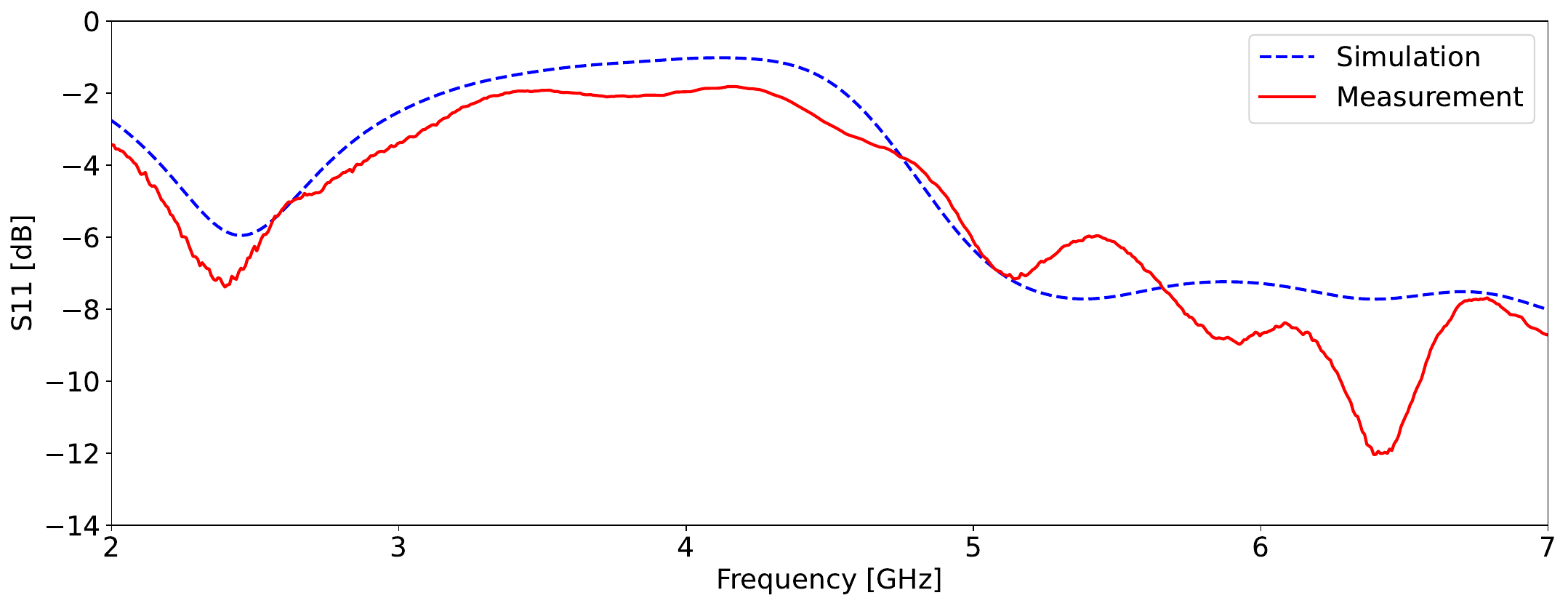}
\label{ex2_s11}}

\subfloat[]{\includegraphics[width=0.8\columnwidth]{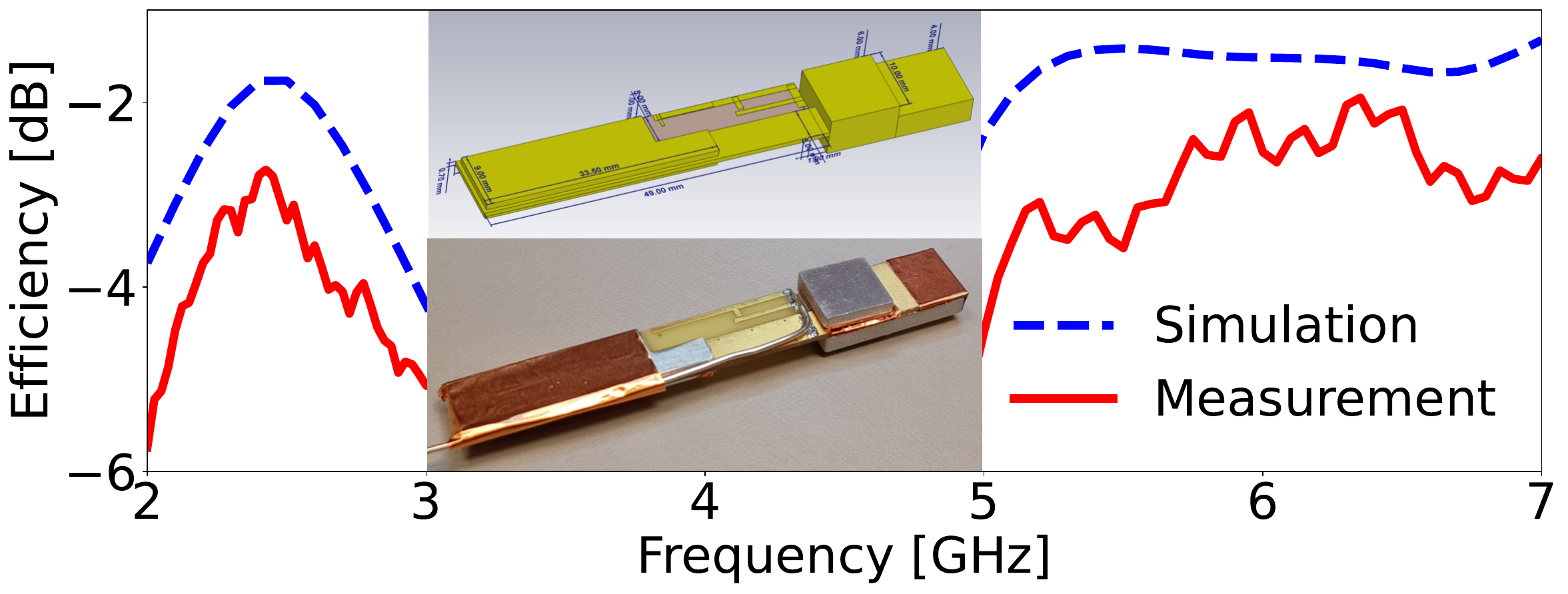}
\label{ex2_eff}}
\caption{Comparison of the final simulation and PCB measurement. (a) Reflection coefficient. (b) Antenna efficiency on the frequency bands of interest.}
\label{ex2_pcb_results}
\end{figure}

\begin{figure}[!t]
\centering
\subfloat[]{\includegraphics[width=0.25\columnwidth]{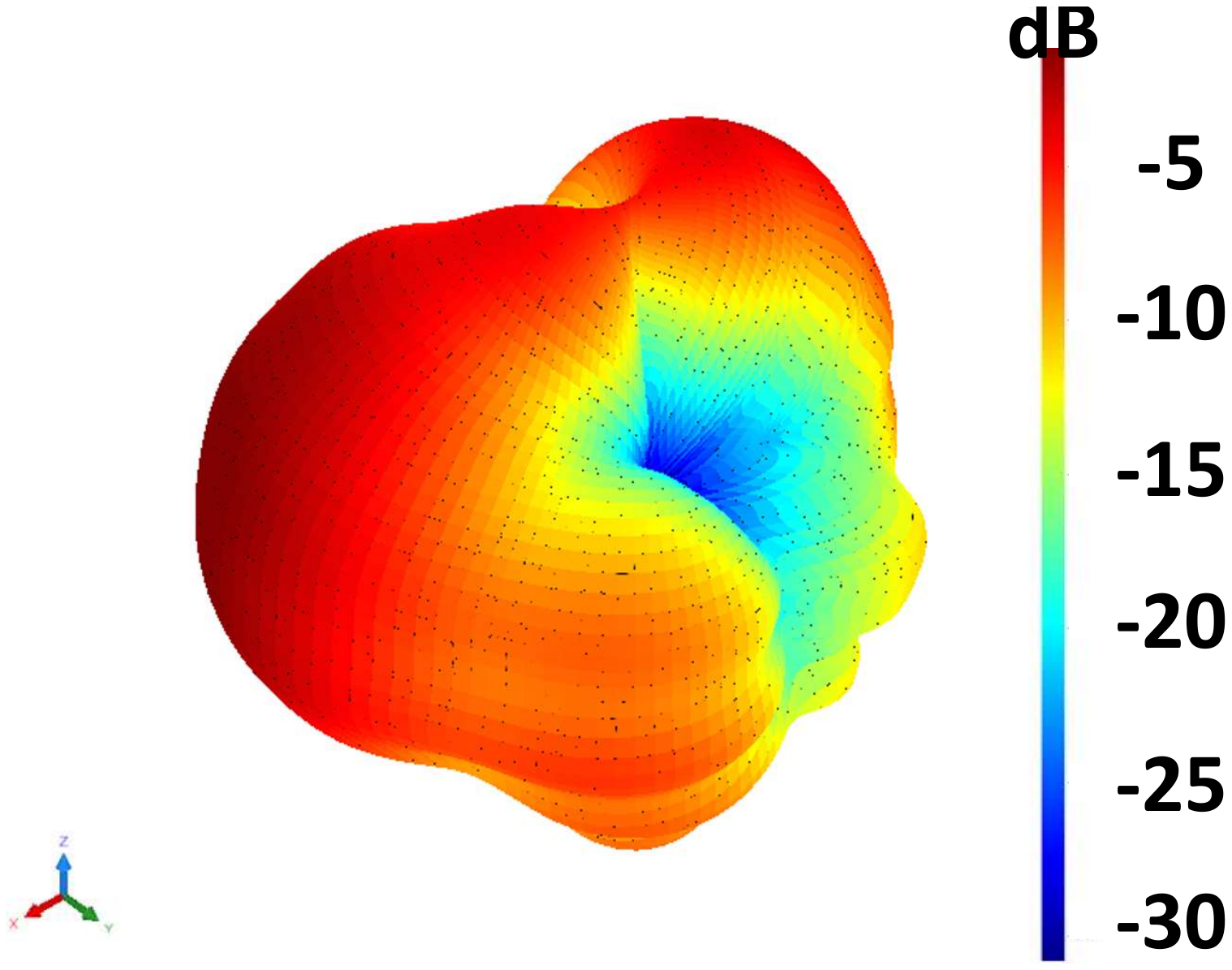}
\label{ex2_gain_2_H}}
\subfloat[]{\includegraphics[width=0.25\columnwidth]{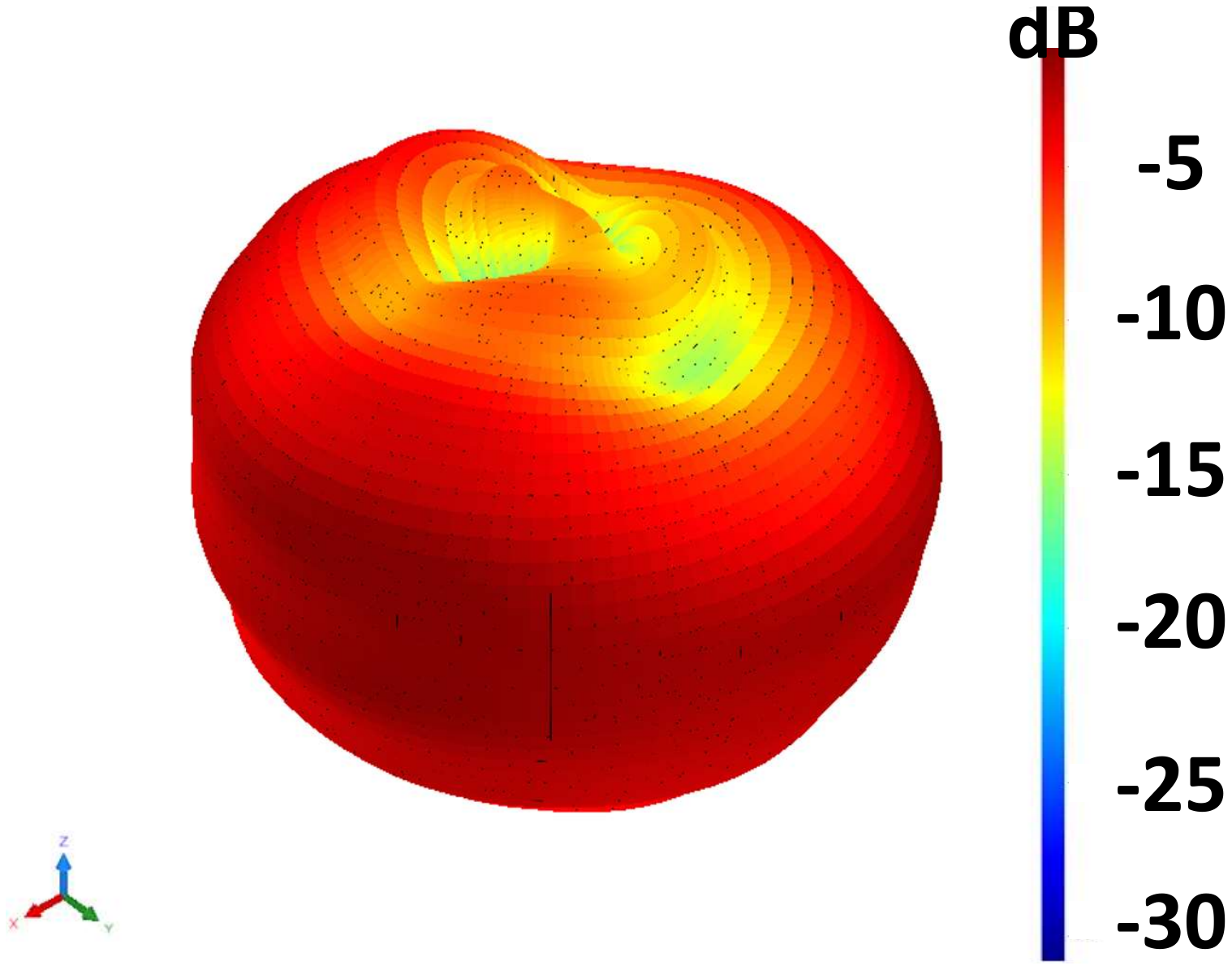}
\label{ex2_gain_2_V}}
\subfloat[]{\includegraphics[width=0.25\columnwidth]{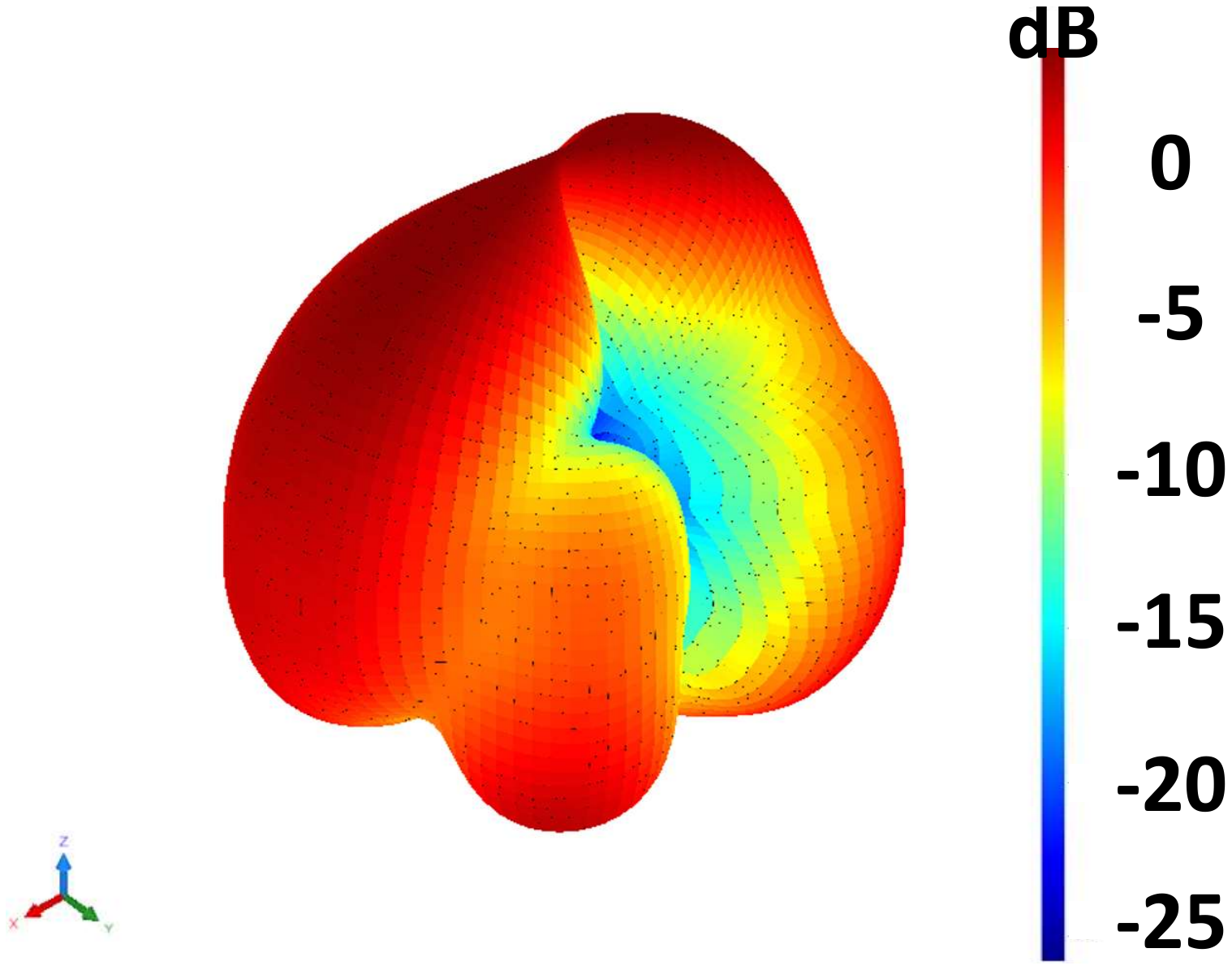}
\label{ex2_gain_6_H}}
\subfloat[]{\includegraphics[width=0.25\columnwidth]{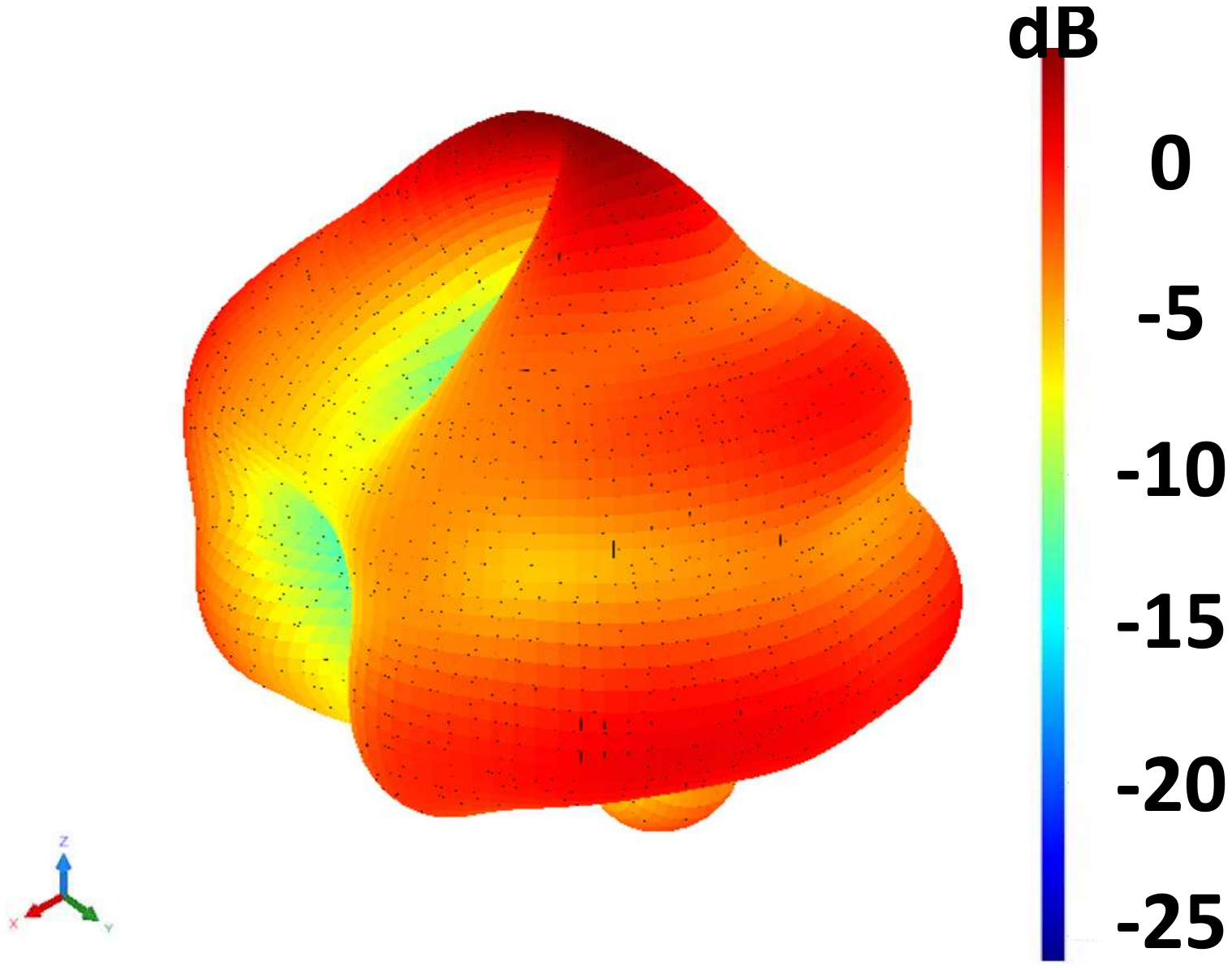}
\label{ex2_gain_6_V}}
\caption{Gain of the final antenna model (a) Horizontal polarization E-field at 2.4 GHz. (b) Vertical polarization E-field at 2.4 GHz. (c) Horizontal polarization E-field at 6 GHz. (d) Vertical polarization E-field at 6 GHz. }
\label{ex2_gain}
\end{figure}

The reflection coefficient and efficiency of this PCB antenna are shown in Fig.~\ref{ex2_pcb_results}. The antenna gain at 2.4 GHz and 6 GHz are visualized in Fig.~\ref{ex2_gain}.

\subsection{Post-Optimization Analysis}
\label{analysis}

So far only five sets of dimensions have been examined in the dimension selector, and we will investigate more in our future research work. Before that, we use the post-optimization dimensions, as reported in Table~\ref{table:e2_opt}, to elaborate the feasibility of selecting dimensions based on the random-position sampling statistics. The dimension set of $6^*$ is the dimensions of the final design that meets the goal. We also treat it as a candidate in the dimension selector and the statistics of 100 samples with random positions are shown in Fig.~\ref{ex2_dim}. Furthermore, we round values in the set $6*$ into integers to get the set $6'$ in Table~\ref{table:e2}. The two dimension sets are significantly better than the original five candidates, suggesting some correlation between the quality of the dimensions and the statistics of the random-position samples.

\section{Conclusions}
\label{conclusion}

With two examples, the proposed PCB antenna design workflow and the image based PCB antenna classifier have been demonstrated. It should be noted that the antenna dimensions are intentionally decoupled with the positions in order to reduce the complexity of antenna designs, and ease of AI treatment. Following the proposed methodology, high quality antenna prototypes have been developed without prior antenna design experience. With them, the optimization process is no longer challenging. The image-based classifier in the antenna model generator plays an important role in the development of the prototype. In the second examples, We can see that the generator has learned intelligence and moved the most of the components to the top of the design area, which aligns with the expectation from design experts. In the analysis of post-optimization, we observed the correlation between the qualities of component dimensions and the statistics of the random-position samples. Since the quality of the component dimensions can be judged before considering the positions, the proposal of separating the dimensions and positions should be valid.

\appendix
\section{Rules of Placing Fixed-Dimension Elements}
\label{Appendix}

\begin{figure}[!t]
\centering
\subfloat[]{\includegraphics[width=0.7\columnwidth]{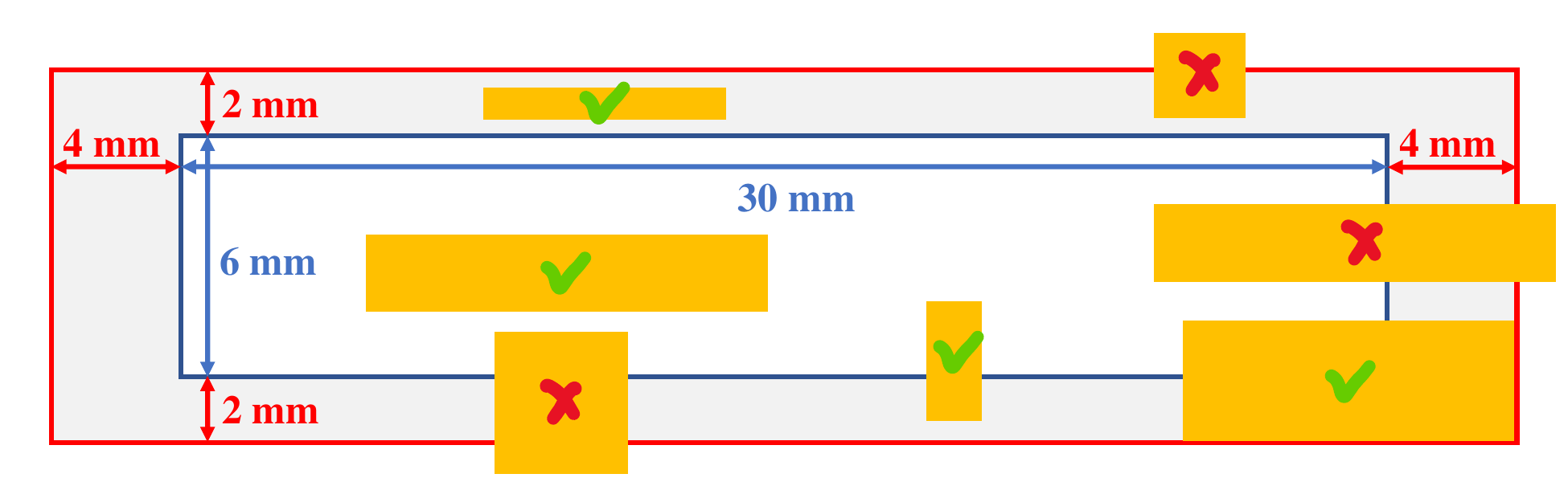}
\label{restrict}}

\subfloat[]{\includegraphics[width=0.7\columnwidth]{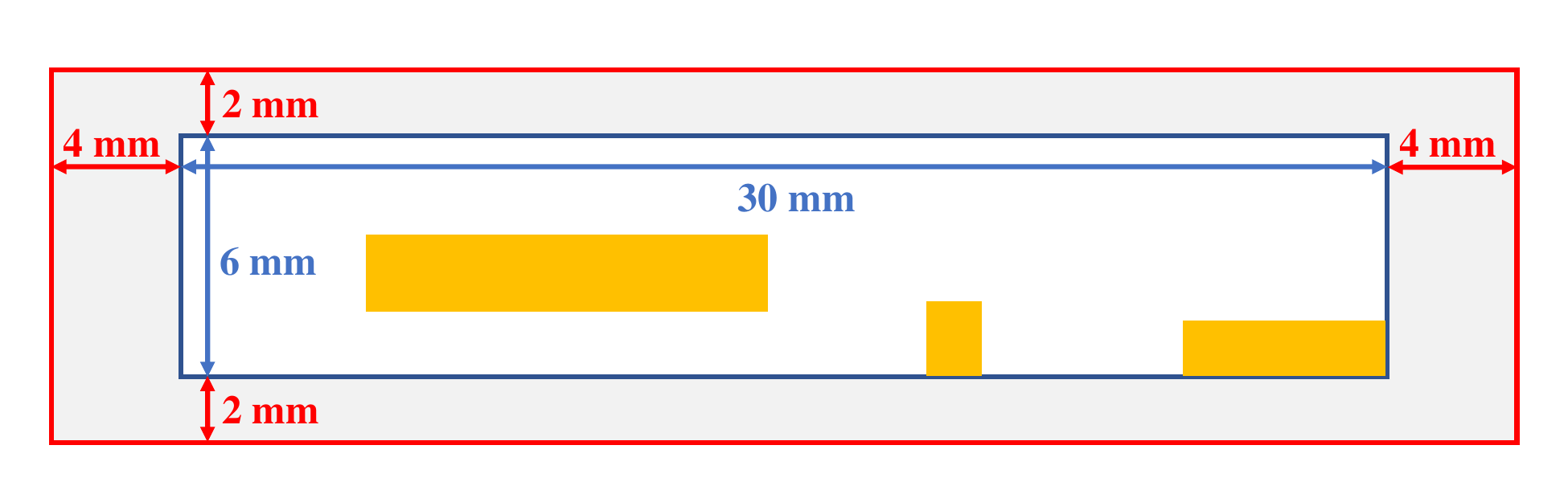}
\label{clipping}}
\caption{Antenna design space and its extension for placing metal components. (a) Position restriction: The component must be in the extended area. (b) Clipping: Only metals within the design space can remain in the simulation.}
\label{restriction}
\end{figure}

We often observe that the conductor part of many PCB antennas consists of a number of basic geometric shapes like rectangles\cite{10237784}. The dimension of each rectangle is parameterized by the width (w) and height (h). The position of each rectangle is parameterized by the x- and y-coordinates of its lower left corner. When position changes, these rectangles can be reassembled into dissimilar antenna types, which enables us to explore opportunistic and inspirational designs. If the fixed-dimension rectangles are tightly squeezed in a compact antenna design area, during the AI design process, the possibility of a random rectangle touching the boundary, such as connecting to the ground (y = 0), becomes negligible. Therefore, it is desirable to reserve an extension room for these rectangles to move around while ensuring them remaining inside. Considering this, AI robot will first disqualify the three rectangles shown in Fig.~\ref{restrict}. Next, it might apply specific rules on some particular rectangles for example adding the antenna port on the first rectangle and anchoring its vertical axis of the first rectangle (y = 0.5 mm). Last, restrict all antenna metal pieces to stay within the assigned design volume. Clipping might be applied to remove the metallic parts that exceed their boundaries on needed basis, as shown in Fig.~\ref{clipping}.

\bibliographystyle{unsrtnat}
\bibliography{references}  






\end{document}